%% file: arxiv.tex
\documentclass{arxiv/research}
\usepackage{amsmath}
\usepackage{amssymb}
\usepackage{mathtools}
\usepackage{amsthm}
\usepackage{latexsym}
\usepackage{bm}
\usepackage{fontawesome5}

\usepackage[utf8]{inputenc}
\usepackage{inconsolata}

\usepackage{array}
\usepackage{makecell}
\usepackage{multirow}
\usepackage{tablefootnote}
\usepackage{adjustbox}

\usepackage{enumitem}
\usepackage{url}
\usepackage{pifont}
\usepackage{wrapfig}
\usepackage{float}

\usepackage{listingsutf8}

\usepackage{titlesec}
\titlespacing*{\paragraph}{0pt}{1.1ex}{1em}

\renewcommand\comment[1]{}

\newcommand{\cmark}{\ding{51}}
\newcommand{\xmark}{\ding{55}}
\newcommand{\project}{CodeAlchemy}

\titleformat*{\paragraph}{\bfseries}%\sffamily}

\BeforeBeginEnvironment{tabular}{\begin{adjustbox}{max width=\textwidth}}
\AfterEndEnvironment{tabular}{\end{adjustbox}}

\lstdefinestyle{promptstyle}{
    basicstyle=\ttfamily\scriptsize,
    breaklines=true,
    frame=single,
    captionpos=t,
    backgroundcolor=\color{gray!10},
    literate=
        {`}{{\textasciigrave}}1
        {"}{\textquotedbl}1
        {"}{\textquotedbl}1
        {'}{'}1
        {'}{'}1
        {—}{---}1
        {–}{--}1
        {→}{->}1
        {≥}{>=}1
        {≤}{<=}1
        {✅}{[GOOD]}5
        {❌}{[BAD]}4
        {…}{...}1
        {≠}{!=}1
        {·}{*}1
        {⌊}{[}1
        {⌋}{]}1
        {∎}{QED}1
        {₀}{\_0}2
        {₁}{\_1}2
        {₂}{\_2}2
        {₃}{\_3}2
        {ₙ}{\_n}2
        {ᵢ}{\_i}2
        {ⱼ}{\_j}2
        {ₖ}{\_k}2
}

\definecolor{prismBg}{HTML}{E4EFF5}
\definecolor{prismFrame}{HTML}{A8C8D8}
\newtcolorbox{takeaway}{
    colback=prismBg,
    colframe=black!50,
    arc=4pt,
    boxrule=0.6pt,
    left=4pt,
    right=4pt,
    top=5pt,
    bottom=5pt,
    before skip=10pt,
    after skip=17pt,
    enlarge left by=1.1cm,
    width=\dimexpr\linewidth-2.2cm\relax,
    before upper={\sloppy},
}

\title{\project: Synthetic Code Rewriting at Scale}

\author[]{Ankit Gupta}
\author[]{Aditya Prasad}
\author[]{Rameswar Panda}

\affiliation[]{MIT-IBM Watson AI Lab, IBM Research}

\abstract{
\input{arxiv/abstract}
}

\metadata[\raisebox{-0.1em}{\includegraphics[height=1.1em]{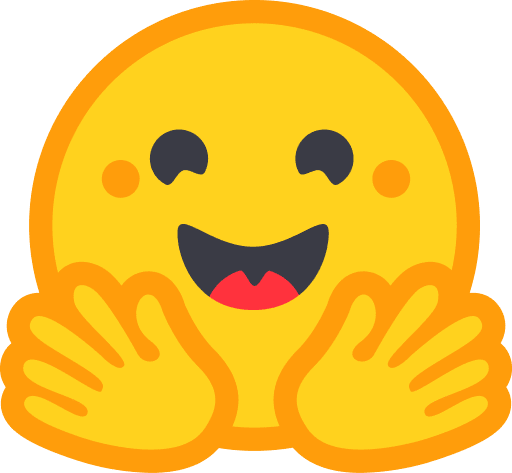}}\ Data]{\href{https://huggingface.co/datasets/open-alchemy/code-alchemy}{huggingface.co/datasets/open-alchemy/code-alchemy}}
\metadata[\faGithub\ Code]{\href{https://github.com/ag1988/code-alchemy}{github.com/ag1988/code-alchemy}}

\begin{document}
\maketitle
{\renewcommand{\thefootnote}{}\footnotetext{Email: \texttt{ankitgupta.iitkanpur@gmail.com}, \texttt{aditya.prasad@ibm.com}, \texttt{rpanda@ibm.com}}}

\input{arxiv/main}

\bibliographystyle{plainnat}
\bibliography{arxiv/refs}

\newpage
\beginappendix
\input{arxiv/appendix}

\end{document}

%% file: arxiv/abstract.tex
Pre-training on raw code teaches syntax but provides sparse signal for diverse real-world task formats. While synthetic data has proven transformative for language models, code remains largely unexplored beyond limited quality improvements. We present \textbf{\project}, a synthetic data generation framework that transforms publicly sourced code into semantically-rich training data through 5 strategies: \textsc{CodeEnhance} (quality-aware rewriting), \textsc{CodeQA} (template-based problems), \textsc{CodeDev} (developer tasks), \textsc{CodeDialogue} (multi-turn conversations), and \textsc{CodeTrace} (execution traces).
We process 3 corpora across \textbf{15} languages to generate \textbf{500B+} tokens of synthetic data plus \textbf{350B} reasoning tokens, orders of magnitude more than prior efforts. \textsc{CodeTrace} instruments and executes \textbf{1.3M+} files across \textbf{14} languages and \textbf{5K} libraries, capturing control flow, state tracking, and library knowledge. We introduce \textsc{DevEval} (developer tasks) and \textsc{TraceEval} (execution prediction) benchmarks; frontier models like Claude Sonnet 4.5 achieve only 5.6\% exact match on \textsc{TraceEval}, revealing critical gaps in semantic understanding.
Our 3B models achieve 83.5\% on HumanEval, 63.2\% on MBPP, 8.09\% win rate on \textsc{DevEval}, and 15.36 ROUGE-2 on \textsc{TraceEval}, outperforming frontier models \textbf{10×} the size including 27B Gemma-3 and 32B Granite-4.0.

%% file: arxiv/main.tex
\section{Introduction}

\begin{table*}[t]
\centering
\caption{\textbf{Comparison of synthetic code data generation approaches.} \project\ provides comprehensive coverage across all dimensions with multi-language support.}
\label{tab:comparison}
\resizebox{\textwidth}{!}{
% \fontsize{10}{12}\selectfont
\begin{tabular}{l|lll}
\toprule
\textbf{Feature} & \textbf{CodeAlchemy} (this work) & \textbf{Nemotron-Pretraining-Code-v2} & \textbf{SwallowCode-v2} \\
\midrule
\multicolumn{4}{l}{\textit{Data Transformation Strategies}} \\
Quality Enhancement/Rewriting & \cmark~(15 langs) & \cmark~(Python) & \cmark~(Python) \\
Template-based QA Generation & \cmark~(7 langs) & \cmark~(11 langs) & \xmark \\
Grounded Developer Tasks & \cmark~(15 langs) & \xmark & \xmark \\
Multi-turn Conversations & \cmark~(15 langs) & \cmark~(Python, C++) & \xmark \\
Code Execution Tracing & \cmark~(14 langs, 1.3M files) & \xmark & \xmark \\
Cross-language Tasks & \cmark~(14 langs) & \cmark~(Python→C++ only) & \xmark \\
Quality-scored Filtering & \cmark~(120M files scored) & \xmark & \xmark \\
\midrule
\multicolumn{4}{l}{\textit{Infrastructure \& Evaluation}} \\
Multi-language Sandbox Execution & \cmark~(14 langs, 5.4K libs) & \xmark & \xmark \\
Difficulty-based Filtering & \cmark & \xmark & \xmark \\
Execution-based Validation & \cmark~(\textsc{CodeQA}, \textsc{CodeTrace}) & \xmark & \xmark \\
New Evaluation Benchmarks & \cmark~(\textsc{DevEval}, \textsc{TraceEval}) & \xmark & \xmark \\
\midrule
\multicolumn{4}{l}{\textit{Scale \& Coverage}} \\
Total Languages & \textbf{15} & 11 & 1 \\
Tokens & \textbf{500B + 350B reasoning} & $\sim$480B (estimate from 1918GB data) & 50B \\
\bottomrule
\end{tabular}
}
\vspace*{-5pt}
\end{table*}

Large language models (LLMs) for code have advanced through scaling on publicly sourced code \cite{stackv2}, but this paradigm faces 3 limitations: (1) high-quality data is scarce and low-quality code harms performance \cite{smollm2}, (2) raw code provides sparse signal for diverse user interactions (debugging, refactoring, explanations), and (3) next-token prediction teaches syntax but not semantics; a model predicting \texttt{for i in range(n):} learns the pattern but not execution behavior, loop values, or termination conditions. LLM-based data rewriting has proven transformative for text models \cite{mainirephrasing}. Kimi K2 rewrites low-quality documents to improve quality and diversity \cite{kimik2}, while Nemotron-CC and Rewire extract structured QA pairs from documents, boosting MMLU scores \cite{nemotroncc,rewire}.

\begin{figure*}[t]
  \centering  
  \includegraphics[width=1.01\textwidth]{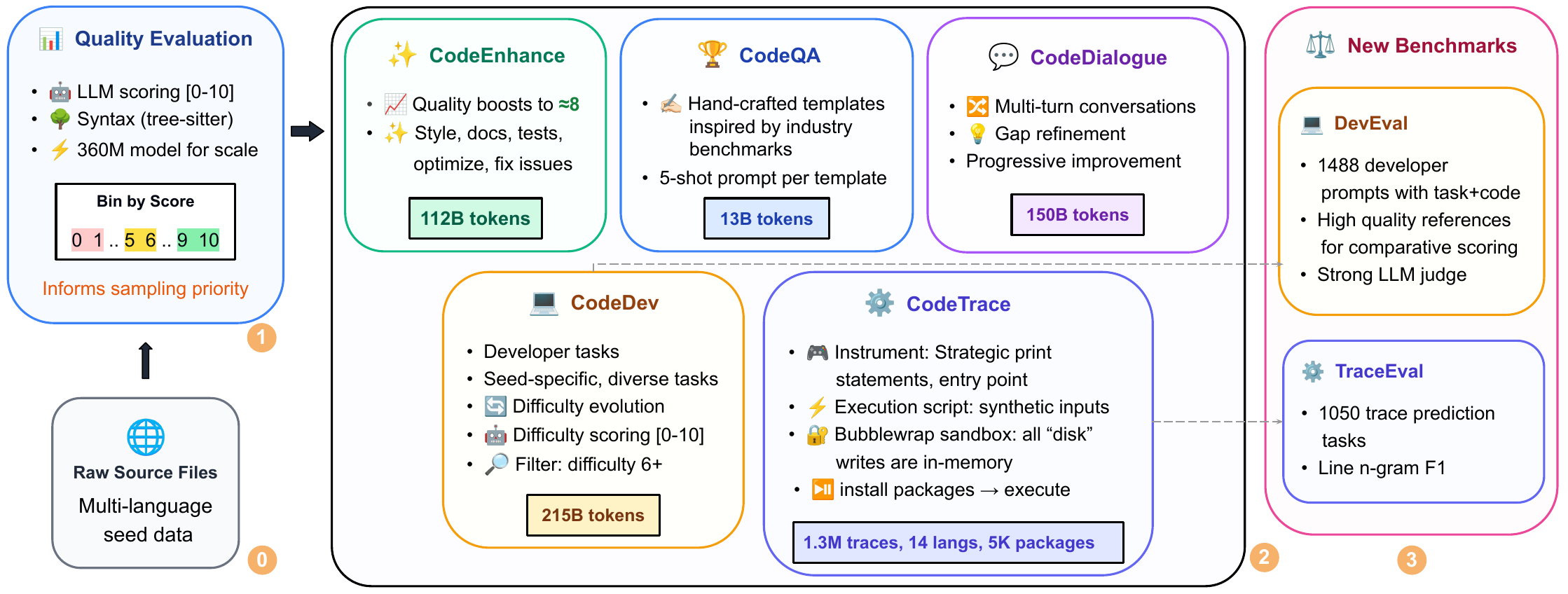}
  \caption{{\bf \project\ Data Pipeline.} Starting from multi-language seed data (Stage 0), we evaluate code quality via LLM scoring (Stage 1), which informs sampling for five synthesis methods (Stage 2): \textsc{CodeEnhance} improves quality, \textsc{CodeQA} uses benchmark-inspired templates, \textsc{CodeDev} generates developer tasks, \textsc{CodeDialogue} creates multi-turn developer-assistant conversations, and \textsc{CodeTrace} produces execution traces from 1.3M runs. The pipeline generates 500B tokens plus 350B reasoning tokens, evaluated on two new benchmarks (Stage 3).}
  \label{fig:pipeline}
  \vspace*{-5pt}
\end{figure*}

Despite these successes, \emph{synthetic code data for pretraining} remains under-explored. SwallowCode focuses exclusively on Python quality enhancement \cite{swallowcode}, while Nemotron-Pretraining-Code-v2 provides QA generation across 11 languages but limits quality rewriting to Python and cross-language tasks to Python→C++ only \cite{nvidia_nemotron_nano_v3_2025} (detailed comparison in Table~\ref{tab:comparison}).
This leaves critical gaps: (1) \textit{Quality:} most code lacks tests, documentation, and error handling; (2) \textit{Format Alignment:} benchmark formats (docstrings, signatures, test cases) are underrepresented in raw code; (3) \textit{Task Diversity:} raw code doesn't capture realistic developer workflows or multi-turn conversations; (4) \textit{Reasoning:} complex QA instances contain implicit reasoning steps; (5) \textit{Semantics:} next-token prediction teaches syntax but not execution semantics (control flow, state tracking, etc). To address these, we propose \textbf{\project}, a multi-faceted pipeline transforming raw code into diverse, semantically-rich training data (Figure~\ref{fig:pipeline}).

\paragraph{\textsc{CodeEnhance}} Publicly sourced code often contains errors, poor naming, and lacks tests, documentation, and error handling \cite{swallowcode}. We score files 0-10 using an LLM, then selectively rewrite low-quality code ($\leq$ 6) to add: unit tests with edge cases, documentation, stubs/mocks for dependencies, style guide adherence, etc. Quality scores improve to $\sim$8 regardless of original quality (Section~\ref{sec:code-enhance}), yielding 120B tokens across 15 languages.

\paragraph{\textsc{CodeQA}} Benchmarks use specific formats (function signatures, docstrings, test cases) rarely seen in raw code. In the text domain, training on QA pairs extracted from documents significantly boosts MMLU performance \citep{mainirephrasing,nemotroncc,rewire}. To investigate whether similar patterns hold for code, we hand-craft 5 QA templates with reference examples, then prompt an LLM to generate instances per source file that match the template format while being \emph{inspired by} source code patterns. Grounding generation in diverse files naturally introduces variation, yielding 22M novel QA pairs.

\paragraph{\textsc{CodeDev}} Raw code provides sparse signal for realistic developer requests. We prompt an LLM to generate diverse developer tasks grounded in each source file: debugging, refactoring, porting, explanations, etc (Table~\ref{tab:deveval-categories}). Each task references \emph{actual code elements}, ensuring specificity and avoiding artificiality of purely prompt-based generation. This yields 215B tokens of 62M prompt-response pairs across 15 languages, plus 68B reasoning tokens.

\paragraph{\textsc{CodeDialogue}} Real developer interactions require multiple turns to refine requirements, debug issues, and explore alternatives. We extend \textsc{CodeDev} into multi-turn conversations by prompting an LLM to generate follow-up turns that identify gaps in previous responses, introduce constraints, or progress tasks naturally. This produces diverse interaction patterns: clarification requests, iterative refinements, debugging sessions, code reviews, yielding 150B tokens across 31M conversations (3.6 rounds average), plus 271B reasoning tokens.

\paragraph{\textsc{CodeTrace}} Next-token prediction provides sparse signal for semantics. Given \texttt{x = foo(y)}, models learn syntax but not what values \texttt{x} takes, which branches execute, or how state evolves. Publicly sourced code cannot be executed directly due to unresolved imports, missing dependencies, and lack of instrumentation. We build execution tracing infrastructure that: (1) instruments 4M files across 14 languages and 5K libraries to emit structured trace events, (2) generates test inputs, (3) executes in isolated containers, (4) filters non-deterministic code ($\sim$75\%). This yields 1.3M (code, trace) pairs capturing control flow, state evolution, and library behavior. Critically, \emph{frontier models struggle}: Claude Sonnet 4.5 achieves only 30.8\% line bigram F1, validating task difficulty. While execution traces have been used to evaluate state tracking abilities of new architectures, prior work uses only toy grammars or simple Python functions \cite{l0bench,ding2024semcoder,whaticannotexecute}.

\paragraph{New Benchmarks} \project\ framework also reveals a gap in how models are evaluated. Existing benchmarks measure algorithmic problem-solving (HumanEval, MBPP, CodeContests) but not whether a model can handle a developer's concrete goal given a source file, or mentally simulate what code does when it runs. To target these gaps we introduce two benchmarks. \textsc{DevEval} contains 1488 diverse developer tasks across 12 languages (debugging, feature extension, porting, etc), evaluating practical abilities beyond isolated function completion. \textsc{TraceEval} contains 1050 execution prediction tasks across 14 languages requiring models to mentally simulate control flow, state evolution, and data transformations. These benchmarks challenge frontier models: Claude Sonnet 4.5 achieves only 5.6\% exact match on \textsc{TraceEval}, revealing critical gaps in semantic understanding.

We open-source \project\ data and codebase under Apache 2.0 license. All instruction prompts used in our work are included in Appendix~\ref{sec:prompts}  and examples are provided in Appendix~\ref{sec:samples}.

% \begin{takeaway}
% \raggedright
% \textbf{Data:} \url{https://huggingface.co/datasets/open-alchemy/code-alchemy}\\
% \textbf{Code:} \url{https://github.com/ag1988/code-alchemy}
% \end{takeaway}

\section{Method}\label{sec:method}

We construct \project\ from 3 code corpora (stack-edu, the-stack-v2-train-smol-ids, RefineCode) across 15 languages: Java, Python, JavaScript, PHP, C, C++, C\#, Typescript, Shell, Go, Markdown, Ruby, Rust, Swift, and SQL \cite{smollm2,stackv2,refinecode}.

\subsection{Enhancing the quality of raw code}\label{sec:code-enhance}

\paragraph{Quality Scoring} To enable targeted rewrites, we prompted \texttt{gpt-oss-20b}\footnote{reasoning effort set to ``medium'' unless explicitly stated as ``high''.} to score stack-edu files from 0-10 using Prompt~\ref{lst:prompt-codeenhance-scoring} \cite{gptoss}. For faster processing, we uniformly sampled 10K scored files per language (except SQL, Markdown) and finetuned \texttt{SmolLM2-360M} on (code, score) pairs to obtain a fast quality scorer \cite{smollm2}.

Figure~\ref{fig:stack-edu-scores} and Table~\ref{tab:stack_edu_score_distribution} show the score distribution of stack-edu: a significant fraction scores 0, and most languages average below 5, indicating the data is far from production quality. We used \texttt{tree-sitter}\footnote{\url{https://github.com/Goldziher/tree-sitter-language-pack}} to tag syntax errors but, unlike SwallowCode and SeedCoder, avoided syntax-based filtering due to high false positives in SQL, C, C++, and C\# \cite{swallowcode,seedcoder}.

\paragraph{\textsc{CodeEnhance}} We enhanced code quality using \texttt{gpt-oss-20b} with Prompt~\ref{lst:prompt-codeenhance-rewrite} and used specialized prompts for Markdown, SQL, Shell. Due to compute constraints, we focused on bins 4-6, which have highest improvement potential. We processed 45.7M deduplicated files, yielding 112B tokens.
To evaluate effectiveness, we uniformly sampled 500 files per language-quality bin and scored both raw and rewritten versions using \texttt{gpt-oss-120b} as judge. Figure \ref{fig:quality-scoring-heatmap} shows rewriting improves quality to $\sim$8 regardless of original score, with mass concentrated above the diagonal.

\subsection{Template-based QA Generation}\label{sec:code-qa}

Raw code repositories provide limited signal for benchmark formats. While functions may implement sophisticated algorithms, they rarely include formal problem statements, signatures, docstrings, or test cases; creating format mismatch for models trained on raw code.

We manually crafted 5 QA templates inspired by prominent benchmarks (basic programming, competitive programming, function execution, code completion, trace prediction), each with 5-6 reference examples demonstrating desired format, style, and difficulty. Unlike purely prompt-based synthesis that produces repetitive problems, we ground generation in actual source code. For each (source file, template) pair, Prompt~\ref{lst:prompt-codeqa-samples} generates instances that match the template format while drawing inspiration from algorithmic patterns in the source file.

We applied this pipeline to quality-scored Python files (score $\geq$7), generating 2 instances per file-template pair using \texttt{gpt-oss-20b} (high). For competitive programming, we used \texttt{gpt-oss-120b} (high) with separate problem and solution generation steps. Each instance is sandbox-executed to verify correctness, eliminating $\sim$28\% of samples. We translated basic programming samples to 6 additional languages using dedicated templates per language. This yielded the \textsc{CodeQA} dataset: 22M QA pairs spanning 13B tokens across 7 languages, with diverse problem types and difficulty while maintaining benchmark-quality formats.

\subsection{Developer Tasks}\label{sec:code-dev}

For 12.6M stack-edu source files spanning 15 languages, we generate realistic developer prompts and responses.

\noindent\textbf{Prompt Generation.} Using \texttt{gpt-oss-20b} and Prompt~\ref{lst:prompt-codedev-create-prompts}, we generate 10-15 diverse prompts per file. The prompt requires concrete references to code elements and targets 12 task categories (code comprehension, debugging, feature extension, cross-language, etc) with diversity across 7 dimensions (scope, scenario, constraints, audience, format, style, difficulty).

\noindent\textbf{Response Generation.} Each prompt-code pair is answered using \texttt{gpt-oss-20b} with Prompt~\ref{lst:prompt-codedev-response}, instructing expert-level responses, yielding 153M (prompt+code, response) pairs.

\noindent\textbf{Difficulty Evolution.} To increase complexity, we apply Prompt~\ref{lst:prompt-codedev-prompt-evolve} to ``Complex''-tagged prompts using 4 strategies: \textit{Mutation} (constraint stacking, adversarial twists), \textit{Crossover} (fusing prompts into workflows), \textit{Hybrid} (combining both), and \textit{Invention} (generating new prompts). This produces 112M additional pairs: total 265M pairs spanning 685B tokens with 230B reasoning tokens.

\noindent\textbf{Quality Filtering.} To keep the average data quality high, we use \texttt{gpt-oss-20b} (high) with Prompt~\ref{lst:prompt-codedev-prompt-scoring} to score prompts on three 0-9 scales (\textit{validity}, \textit{difficulty}, \textit{training value}), retaining prompts with validity $\geq$8, and training value $\geq$8 or (training value $=$7 and difficulty $\geq$7).
The \textsc{CodeDev} dataset comprises 62M pairs spanning 207B tokens plus 68B reasoning tokens.

\subsection{Developer-Assistant Conversations}\label{sec:code-dialogue}

Real developer interactions require multiple turns to refine requirements and debug issues. We extend \textsc{CodeDev} pairs (with difficulty $\geq$7) into multi-turn conversations using \texttt{gpt-oss-20b} (high).

Starting from a \textsc{CodeDev} pair as the initial exchange, we iteratively generate conversation turns. For each conversation history, Prompt~\ref{lst:prompt-codedialogue-user} generates the next developer turn by identifying gaps or issues in the previous assistant response and building progressively on the discussion, enforcing diverse follow-up types: clarification requests, iterative refinements, debugging, testing, extensions, and code review. Each developer turn is answered using Prompt~\ref{lst:prompt-codedialogue-assistant}, which generates expert responses that address the user's request. Conversations exceeding 100K characters are not extended any further.

This yielded 31M conversations spanning 150B tokens (excluding first-round \textsc{CodeDev} data) and averaging 3.6 rounds, plus 271B reasoning tokens from the assistant response generation

\subsection{Code Tracing}\label{sec:code-trace}

Next-token prediction provides sparse signal for program semantics. Given \texttt{x = foo(y)}, models learn syntactic patterns but not what values \texttt{x} can take, which branches \texttt{foo} executes, or how state evolves. To address this, we build a large-scale execution tracing infrastructure.

\noindent\textbf{Instrumentation.} We use Prompt~\ref{lst:prompt-codetrace-instrument} to transform 4M quality-scored files from stack-edu (score $\geq$7) across 14 languages into deterministic, instrumented programs. The prompt inserts 15-25 trace emission points outputting structured events to \texttt{STDERR} as \texttt{TRACE:<TYPE>:<LOC>:<STATE>}, where \texttt{TYPE} $\in$
\{IN, OUT, VAR, BRANCH, LOOP, ERR, TRANSFORM\}, \texttt{LOC} identifies the function/block/line, and \texttt{STATE} captures program state. The prompt enforces 15 complex trace patterns: checkpoint thresholds (\texttt{sum > checkpoint+500}), stack deltas (\texttt{|len(stack)-prev| > 5}), statistical conditions (\texttt{median(buf)-mean(buf) > std(buf)}), etc designed to challenge semantic understanding rather than test memorization. Only popular mainstream libraries are retained for executability.

\noindent\textbf{Test Input Generation.} For each instrumented file, Prompt~\ref{lst:prompt-codetrace-test} generates a bash script with 3-5 tests producing structurally distinct traces via CLI arguments, stdin, or heredocs. The prompt prescribes 10K-15K stderr characters to keep traces predictable.

\noindent\textbf{Execution.} We execute instrumented files and test scripts in isolated sandboxes that install dependencies on the fly through standard package managers and capture structured trace events from \texttt{STDERR} (sandbox details in Appendix~\ref{sec:sandbox}). To filter non-deterministic code, each file is executed thrice with identical seeds. Files producing empty or inconsistent traces are removed ($\sim$75\% filtered).
This constitutes \textsc{CodeTrace}: 1.3M (code, trace) pairs across 5430 libraries, 14 languages, and 7.4B tokens. Traces capture control flow (BRANCH, LOOP), state evolution (VAR), function boundaries (IN, OUT), data transformations (TRANSFORM), error conditions (ERR), and library API behavior, providing rich signal for teaching program execution, not just syntax.

\subsection{Evaluating Developer Knowledge}\label{sec:dev-eval}

\begin{table}[t]
\caption{\textbf{Performance of candidate models on \textsc{DevEval}} using Claude 4.5 Sonnet (max 50K thinking tokens) as reference and gpt-oss-20b (high) as judge. reasoning\_effort=high for all models.}
\label{tab:deveval-results}
\fontsize{9.5}{11.5}\selectfont
\centering
\begin{tabular}{lcccc}
\toprule
\textbf{Model} & \makecell{\textbf{Params} \\ \textbf{(B)}} & \textbf{Win \%} & \makecell{\textbf{Candidate} \\ \textbf{score}} & \makecell{\textbf{Reference} \\ \textbf{score}} \\
\midrule
gpt-oss-120b (high) & 117  & 61 & 8.82 & 8.24 \\
gpt-oss-20b (high)  & 20   & 39 & 8.36 & 8.48 \\
GLM-5.1             & 754  & 32 & 8.17 & 8.67 \\
Qwen3.6-35B-A3B     & 35   & 22 & 7.59 & 8.75 \\
MiniMax-M2.5        & 230  & 17 & 6.82 & 8.86 \\
gemma-4-E2b-it      & 5.1  &  3 & 5.99 & 9.11 \\
granite-4.0-h-small & 32   &  2 & 5.30 & 9.15 \\
\bottomrule
\end{tabular}
\end{table}

To evaluate LLMs on developer tasks, we created \textsc{DevEval} using the \textsc{CodeDev} pipeline (Section~\ref{sec:code-dev}) on 150K RefineCode files not in stack-edu. We generated 1.8M prompts using \texttt{gpt-oss-20b} (high), retaining only ``Complex'' prompts with difficulty $\geq$ 6.

\noindent\textbf{Ensuring Diversity.} Following \cite{arenalearning}, we embedded prompts (without the source code) using \texttt{embeddinggemma-300m} and clustered into 100 clusters per language-difficulty bin \cite{embeddinggemma}. For each language, we sampled 41 hardest samples round-robin across clusters (starting from bin 9→6), then 30 samples each from bins 6-8 (round-robin across clusters), yielding 1488 prompts across 12 languages. The prominent task categories are shown in Table~\ref{tab:deveval-categories}.

\noindent\textbf{Evaluation Protocol.} We generated reference responses using Claude 4.5 Sonnet (50K thinking tokens), then evaluated model responses via preference scoring with \texttt{gpt-oss-20b} (high) as judge (Prompt~\ref{lst:prompt-deveval}). We also experimented with using Claude 4.5 Sonnet itself as the judge but it did not change the rankings. Each comparison was performed twice with switched order to mitigate positional bias \cite{codearenaeval}. Samples scoring higher than the reference count as wins (0.5 for ties). Performance of models on \textsc{DevEval} is included in Table~\ref{tab:deveval-results}.

\begin{table}[t]
\caption{\textbf{Performance on \textsc{TraceEval}}. Low performance of gpt-oss-20b (high) is due to reasoning length exceeding the maximum model length resulting in empty final response.}
\label{tab:traceeval-results}
\centering
\fontsize{9.5}{11.5}\selectfont
\begin{tabular}{lcc}
\toprule
\textbf{Model} & \makecell{\textbf{Exact} \\ \textbf{Match (\%)}} & \makecell{\textbf{ROUGE-2} \\ \textbf{(\%)}} \\
\midrule
gemma-3-27b-it & 0.4 & 9.5 \\
gpt-oss-20b (medium) & 1.4 & 17.0 \\
gpt-oss-20b (high) & 1.0 & 8.9 \\
\makecell[l]{max(gpt-oss-20b medium, \\ gpt-oss-20b high)} & 1.9 & 19.9 \\
Claude 4.5 Sonnet (tools off) & 5.6 & 30.8 \\
\bottomrule
\end{tabular}
\end{table}

\subsection{Evaluating Mental Execution Abilities of LLMs}\label{sec:trace-eval}

To evaluate LLMs' mental execution abilities, we created \textsc{TraceEval} using the \textsc{CodeTrace} pipeline (Section \ref{sec:code-trace}) on held-out stack-edu files, generating 94K candidate tasks. To ensure challenging but predictable traces, we applied multiple quality filters.

\noindent\textbf{Filtering for Predictability.} We removed samples requiring external packages, then used Prompt~\ref{lst:prompt-traceeval-unpredictable} with \texttt{gpt-oss-20b} (high) to filter unpredictable elements (non-determinism, runtime-dependent values, external state) and computationally intensive operations (cryptographic hashes, seeded PRNGs). Moreover, we used Prompt~\ref{lst:prompt-traceeval-clean-trace} to remove traces with system-generated noise (compiler warnings, errors, deprecations) and removed samples exceeding 300 \texttt{TRACE:} lines.

\noindent\textbf{Difficulty Calibration.} We evaluated \texttt{gpt-oss-20b}, \texttt{gpt-oss-20b} (high), and \texttt{gemma-3-27b-it} on remaining samples using line-level ROUGE-2 F1 scores (bigram overlap between predicted and ground truth traces) \cite{lin-rouge}. Samples where all models scored < 5\% were removed as excessively difficult. For 14 languages, we selected the 75 hardest samples per language (lowest maximum score across models), yielding 1050 samples comprising \textsc{TraceEval}. Performance of models on \textsc{TraceEval} is included in Table~\ref{tab:traceeval-results} with frontier models like Claude Sonnet 4.5 scoring only 5.6\% exact match.

\section{Experiments}\label{sec:experiments}

We systematically investigate the utility of each constituent of \project\ by investigating 10 Research Questions (RQs) that can be grouped into three questions: (1) is each component necessary and complementary? (RQ1-2); (2) how should components be combined, and how much data is needed? (RQ3-5); (3) how does \project\ stand relative to other open alternatives? (RQ6-10). As our default setup, we perform continual pretraining of a 3B parameter base checkpoint trained on 12 trillion tokens from a mixture of natural language, code, academic text, math and multilingual data. In each RQ, we differ only the data mixtures holding all other hyperparameters constant. Evaluation is performed on standard coding benchmarks such as HumanEval, HumanEval+, MBPP, MBPP+, CruxEval, MultiPL-E as well as our new benchmarks \textsc{DevEval} and \textsc{TraceEval} which we introduce in this work \cite{humaneval,evalplus,mbpp,cruxeval,multipl-e}. Training and evaluation hyperparameters are included in Appendix~\ref{sec:hyperparams}.

\paragraph{RQ1: Does quality-enhanced data make raw code redundant?} We anneal the base model on 100B tokens using three mixtures: (1) only raw code, (2) only \textsc{CodeEnhance}, and (3) 50-50 mixture of both. As shown in Table~\ref{tab:results-main}, \textsc{CodeEnhance} achieves substantial gains on HumanEval (40.9 vs 28.7 for raw) but suffers dramatic performance drops on MBPP (5.6 vs 58.7). The 50-50 mixture preserves MBPP performance (60.3) while maintaining modest HumanEval scores (30.5). This suggests that standardization in \textsc{CodeEnhance}, due to adherence to official style guides, helps on HumanEval but reduces the diversity needed for MBPP, which contains varied documentation styles and naming conventions. Adding \textsc{CodeQA} data (without the contest problems) to either mixture further improves the balance, with \textsc{CodeEnhance} + \textsc{CodeQA}-subset achieving the best HumanEval score (51.2) while partially recovering MBPP scores (33.3). As reported by \citet{swallowcode}, MBPP contains functions with non-standard mixed-case names (e.g. \texttt{is\_Power\_Of\_Two}) that violate PEP 8; \textsc{CodeEnhance} rewrites these to snake\_case, causing errors at evaluation time when the harness calls the original name. This drop reflects a naming convention mismatch rather than a genuine capability gap - evidenced by our best model achieving strong scores on both MBPP (63.2) and MBPP+ (53.4) once data diversity is restored (RQ4/RQ5).
\begin{takeaway}
Quality-enhanced data does not make raw code redundant; diversity in style and conventions is essential for robust model performance across benchmarks.
\end{takeaway}

\definecolor{highlight}{RGB}{255,255,200}

\begin{table*}[t]
  \centering
  \caption{\textbf{Model performance for different experiments in Section~\ref{sec:experiments}}. Benchmarks: HE = HumanEval, HE+ = HumanEval+, MB = MBPP, MB+ = MBPP+, CX-I/O = CruxEval Input/Output, MLE = MultiPL-E, DE = \textsc{DevEval} (Win \%), TE = \textsc{TraceEval} (ROUGE-2). Second best scores are underlined. Mix1, Mix2, Mix3 are defined in Table~\ref{tab:rq3-mixtures}.}
  \label{tab:results-main}
  {\fontsize{8.6}{10.6}\selectfont
  \setlength{\tabcolsep}{3pt} 
  \begin{tabular}{ll|cc|cc|cc|ccccc|cc}
    \toprule
    \textbf{Model} & \makecell[c]{\textbf{Tokens}\\ \textbf{(B)}} & \textbf{HE} & \textbf{HE+} & \textbf{MB} & \textbf{MB+} & \textbf{CX-I} & \textbf{CX-O} & \makecell{\textbf{MLE}\\\textbf{cpp}} & \makecell{\textbf{MLE}\\\textbf{js}} & \makecell{\textbf{MLE}\\\textbf{go}} & \makecell{\textbf{MLE}\\\textbf{java}} & \makecell{\textbf{MLE}\\\textbf{sh}} & \textbf{DE} & \textbf{TE} \\
    \midrule
    \multicolumn{15}{l}{\textbf{RQ1}} \\
    Raw & 100 & 28.7 & 25.6 & 58.7 & 46.8 & 31.9 & 33.4 & 29.5 & \underline{26.1} & 74.7 & \textbf{20.7} & \textbf{9.7} & 0 & 0.44 \\
    \textsc{CodeEnhance} & 100 & \underline{40.9} & \underline{39.6} & 5.6 & 5.0 & \textbf{37.8} & \underline{36.7} & 23.1 & 7.1 & 63.5 & 3.7 & 4.4 & 0.07 & 0.35 \\
    \makecell[l]{50\% Raw + 50\% \textsc{CodeEnhance}} & 100 & 30.5 & 26.8 & \underline{60.3} & \underline{49.5} & \underline{36.4} & 36.0 & \textbf{31.4} & \textbf{28.0} & \underline{75.7} & 3.8 & 6.3 & 0.03 & 0.45 \\
    \makecell[l]{Raw + \textsc{CodeQA}-subset} & 100 & 29.9 & 26.2 & \textbf{61.4} & \textbf{50.8} & 31.5 & 33.7 & \underline{30.0} & \textbf{28.0} & \textbf{76.7} & \underline{18.8} & \underline{7.4} & 0 & 0.43 \\
    \makecell[l]{\textsc{CodeEnhance} + \textsc{CodeQA}-subset} & 100 & \textbf{51.2} & \textbf{48.2} & 33.3 & 27.0 & \textbf{37.8} & \textbf{38.1} & 20.5 & 12.6 & 74.0 & 2.9 & 3.7 & 0.07 & 0.47 \\
    \midrule
    \multicolumn{15}{l}{\textbf{RQ2}} \\
    \textsc{CodeEnhance} & 125 & \underline{41.5} & \textbf{39.6} & 11.9 & 10.6 & \underline{35.8} & \underline{37.9} & 28.4 & 13.9 & \underline{60.0} & 2.9 & \textbf{8.0} & 0.00 & 0.46 \\
    \textsc{CodeTrace} & 10 & 18.9 & 17.7 & 0.8 & 0.8 & 29.3 & 31.6 & 5.8 & 8.6 & 27.8 & 8.0 & \underline{7.0} & 0.00 & \textbf{13.55} \\
    \textsc{CodeQA} & 10 & \textbf{42.7} & \underline{38.4} & 34.7 & 28.3 & 33.0 & 32.6 & 14.9 & \textbf{40.2} & \textbf{76.6} & 13.0 & 0.0 & 0.30 & 0.37 \\
    \textsc{CodeDev} & 200 & 32.3 & 28.7 & \underline{41.3} & \textbf{36.0} & 33.6 & 36.9 & \underline{31.8} & \underline{31.8} & 47.7 & \underline{23.0} & 6.0 & \textbf{6.73} & 0.71 \\
    \textsc{CodeDialogue} & 250 & 40.2 & 36.6 & \textbf{46.6} & \underline{34.4} & \textbf{37.6} & \textbf{38.3} & \textbf{37.0} & 26.7 & 62.0 & \textbf{23.6} & 6.6 & \underline{6.25} & \underline{1.10} \\
    \midrule
    \multicolumn{15}{l}{\textbf{RQ3}} \\
Raw & 100 & 28.7 & 25.6 & \textbf{58.7} & \textbf{46.8} & 31.9 & 33.4 & 29.5 & 26.1 & \textbf{74.7} & 20.7 & \textbf{9.7} & 0.00 & 0.44 \\
Mix1 & 100 & \underline{40.2} & 34.1 & 38.6 & 30.2 & \underline{37.7} & \textbf{40.8} & 40.9 & \underline{55.8} & 53.6 & \underline{32.6} & 5.0 & \textbf{6.50} & 12.10 \\
Mix1 (without \textsc{CodeTrace}) & 100 & \textbf{42.7} & \textbf{38.4} & 31.0 & 24.9 & \textbf{37.8} & 38.5 & 37.6 & \textbf{55.9} & 17.8 & 21.6 & 5.0 & 4.67 & 0.75 \\
Mix2 & 100 & 39.6 & 31.7 & 45.2 & 37.6 & 33.2 & 40.1 & \textbf{46.0} & 52.5 & 71.6 & 30.4 & 4.5 & 5.42 & \textbf{12.42} \\
Mix3 & 100 & \underline{40.2} & \underline{36.6} & \underline{52.9} & \underline{45.0} & 35.8 & \underline{40.2} & \underline{44.3} & 52.0 & \underline{73.7} & \textbf{33.3} & \underline{6.1} & \underline{5.46} & \underline{12.19} \\
    \midrule
    \multicolumn{15}{l}{\textbf{RQ4}} \\
    Mix1 & 600 & \textbf{46.3} & \textbf{40.2} & \underline{43.9} & \underline{33.9} & \textbf{40.7} & \textbf{44.3} & \textbf{37.1} & \textbf{56.1} & \underline{72.1} & \textbf{32.5} & \textbf{10.1} & \textbf{8.09} & \textbf{15.36} \\
    \textsc{CodeEnhance} & 600 & \underline{35.4} & \underline{32.3} & 4.0 & 3.4 & \underline{36.4} & \underline{38.9} & 25.5 & 14.8 & 60.3 & 4.5 & \underline{9.4} & \underline{0.10} & 0.37 \\
    Raw & 600 & 31.1 & 28.0 & \textbf{52.9} & \textbf{43.4} & 34.8 & 34.0 & \underline{30.4} & \underline{28.5} & \textbf{72.2} & \underline{27.8} & 7.5 & 0.00 & \underline{0.38} \\
    \midrule
    \multicolumn{15}{l}{\textbf{RQ5}} \\
    \textsc{CodeDev} & 200 & \underline{32.3} & \underline{28.7} & \textbf{41.3} & \textbf{36.0} & \underline{33.6} & \underline{36.9} & \underline{31.8} & \underline{31.8} & \underline{46.8} & \underline{0.0} & \underline{6.0} & \underline{6.73} & \textbf{0.71} \\
    \makecell[l]{\textsc{CodeDev}+reasoning→\textsc{CodeDev}} & \makecell[l]{150\\+50} & \textbf{36.0} & \textbf{31.7} & \underline{39.7} & \underline{33.6} & \textbf{35.9} & \textbf{39.3} & \textbf{33.0} & \textbf{34.9} & \textbf{59.9} & \textbf{30.5} & \textbf{8.9} & \textbf{7.14} & \underline{0.67} \\
    % \midrule
%     \multicolumn{15}{l}{\textbf{RQ6\ \ \colorbox{highlight}{(Python only)}}} \\
% \makecell[l]{Raw + SFT on OpenCodeInstruct} & & \underline{76.8} & \underline{70.7} & \underline{57.4} & \underline{49.2} & \underline{30.7} & \underline{29.3} &  &  &  &  & &  & \\
% \makecell[l]{Mix1 + SFT on OpenCodeInstruct} & & \textbf{83.5} & \textbf{78.7} & \textbf{63.2} & \textbf{53.4} & \textbf{37.0} & \textbf{39.2} &  &  &  &  &  &  &  \\
    % \midrule
    % \multicolumn{15}{l}{\textbf{RQ7\ \ \colorbox{highlight}{(Python only)}}} \\
    % \textsc{CodeEnhance} & 10 & 34.1 & 31.7 & 5.6 & 5.0 & \underline{35.3} & 33.2 & & & & & & & \\
    % SwallowCode & 10 & 43.9 & 39.0 & 43.4 & \underline{36.8} & 31.2 & 31.3 & & & & & & & \\
    % Nemotron-RW & 10 & \textbf{48.2} & \textbf{42.1} & \textbf{64.0} & \textbf{46.6} & 32.3 & 31.9 & & & & & & & \\
    % \makecell[l]{\textsc{CodeEnhance}\\+Nemotron-RW+SwallowCode} & 10 & 45.1 & \underline{41.5} & 19.0 & 13.8 & 34.6 & 33.8 & & & & & & & \\
    % \makecell[l]{50\% \textsc{CodeEnhance}\\+ 50\% Nemotron-RW} & 10 & \underline{45.7} & 40.9 & \underline{58.2} & \underline{39.9} & \textbf{35.7} & 33.2 & & & & & & & \\
    % \textsc{CodeEnhance}+\textsc{CodeQA} & 10 & 45.1 & \underline{41.5} & 16.4 & 13.0 & 33.6 & \textbf{36.4} & & & & & & & \\
    % \makecell[l]{95\% \textsc{CodeEnhance}+\textsc{CodeQA} \\+ 5\% Raw} & 10 & 42.7 & 38.4 & 36.0 & 28.0 & 33.6 & 35.1 & & & & & & & \\
    % \makecell[l]{90\% \textsc{CodeEnhance}+\textsc{CodeQA}\\+ 10\% Raw} & 10 & 37.8 & 34.2 & 48.2 & 39.7 & 34.5 & \underline{35.2} & & & & & & & \\
    \midrule
    \multicolumn{15}{l}{\textbf{RQ8\ \ \colorbox{highlight}{(Python only)}}} \\
    Raw $\leq$ 4 & 10 & 26.2 & 23.2 & \underline{52.6} & \textbf{43.7} & 31.1 & 30.1 & & & & & & & \\
    \textsc{CodeEnhance} $\leq$ 4 & 10 & \underline{36.6} & \underline{33.5} & 2.1 & 1.6 & \textbf{36.1} & \underline{33.6} & & & & & & & \\
    Raw $>$ 4 & 10 & 29.9 & \underline{26.8} & \textbf{53.4} & \textbf{43.7} & 30.0 & 31.2 & & & & & & & \\
    \textsc{CodeEnhance} $>$ 4 & 10 & \textbf{38.4} & \textbf{34.8} & \underline{2.6} & \underline{2.6} & \underline{34.3} & \textbf{34.0} & & & & & & & \\
    % \midrule
    % \multicolumn{15}{l}{\textbf{RQ9\ \ \colorbox{highlight}{(Python only)}}} \\
    % \textsc{CodeEnhance} & 10 & 34.2 & 31.7 & 5.6 & 5.0 & \textbf{35.3} & 33.2 & & & & & & & \\
    % \textsc{CodeEnhance}-gemma-4b & 10 & \textbf{38.4} & \textbf{33.5} & \textbf{63.0} & \textbf{52.1} & \underline{33.8} & \textbf{34.5} & & & & & & & \\
    % \textsc{CodeEnhance}-gemma-1b & 10 & \underline{32.9} & \underline{30.5} & \underline{57.7} & \underline{46.3} & 29.5 & \underline{30.9} & & & & & & & \\
    \bottomrule
  \end{tabular}
  }
\vspace*{-10pt}
\end{table*}

\paragraph{RQ2: Do individual constituents of \project\ provide complementary benefits?} We anneal the base model on individual subsets of \project\ for 1 epoch each without accounting for size differences. As shown in Table~\ref{tab:results-main}, each constituent exhibits distinct strengths. \textsc{CodeTrace} (10B tokens) dramatically outperforms all others on \textsc{TraceEval} (13.55 ROUGE-2); even outperforming a much larger 27B \texttt{gemma-3-27b-it}  (Table~\ref{tab:traceeval-results}). \textsc{CodeQA} achieves strong HumanEval performance (42.7) and partial MBPP recovery (34.7), suggesting its QA format bridges the gap between code enhancement and diverse coding patterns. \textsc{CodeDev} and \textsc{CodeDialogue} lead on developer task benchmarks with \textsc{DevEval} scores of 6.73\% and 6.25\% respectively; even outperforming a much larger 32B \texttt{granite-4.0-h-small} (Table~\ref{tab:deveval-results}). Notably, \textsc{CodeEnhance} alone (125B tokens) continues to exhibit the MBPP degradation pattern (11.9), reinforcing findings from RQ1 that standardization trades diversity for style consistency.
\begin{takeaway}
\project\ constituents provide complementary benefits: \textsc{CodeTrace} for execution reasoning, \textsc{CodeQA} for bridging enhancement and diversity, and \textsc{CodeDev}/\textsc{CodeDialogue} for developer tasks.
\end{takeaway}

\paragraph{RQ3: Does mixing diverse types of data yield balanced performance across benchmarks?} We train on 3 mixtures of \project\ constituents at 100B tokens each, progressively increasing raw code from 0\% (Mix1) to 10\% (Mix2) to 20\% (Mix3) (see Table~\ref{tab:rq3-mixtures}). As shown in Table~\ref{tab:results-main}, all mixtures substantially outperform raw-only training on HumanEval ($\sim$40 vs 28.7), demonstrating that quality-enhanced and specialized data provide clear benefits. However, MBPP performance reveals a trade-off: Mix1 with no raw code achieves only 38.6, while Mix3 with 20\% raw code recovers to 52.9, approaching the raw-only baseline of 58.7. This mirrors the findings from RQ1 that raw code diversity is essential for MBPP. All mixtures achieve strong \textsc{DevEval} and \textsc{TraceEval} scores compared to individual constituents (RQ2), demonstrating that combining diverse data types improves coverage across benchmarks. Mix3 provides the most balanced performance, maintaining HumanEval gains (40.2) while best preserving MBPP scores (52.9) and achieving competitive multilingual results (73.7 on Go, 44.3 on C++).
\begin{takeaway}
Mixing diverse data types retains the strengths of individual constituents, enabling balanced performance across all benchmarks.
\end{takeaway}

\paragraph{RQ4: Does more training budget help and does data repetition across epochs hurt performance?} We compare training with 600B tokens using 3 approaches: 5 epochs over \textsc{CodeEnhance} (112B tokens), 5 epochs over raw code (120B tokens), and 1 epoch over Mix1 (590B tokens). As shown in Table~\ref{tab:results-main}, Mix1 trained for 1 epoch substantially outperforms both \textsc{CodeEnhance} and raw code trained for 5 epochs on HumanEval (46.3 vs 35.4 and 31.1), demonstrating that diversity in data types and rephrasing styles is more valuable than repeated exposure to homogeneous data. The MBPP results further support this finding: while raw code at 5 epochs achieves the highest MBPP score (52.9), Mix1 maintains competitive performance (43.9) despite seeing each example only once. Notably, \textsc{CodeEnhance} at 5 epochs shows severe MBPP degradation (4.0), suggesting that repeated exposure to standardized code exacerbates the diversity loss observed in RQ1 and RQ3. Mix1 also achieves the strongest \textsc{DevEval} (8.09) and \textsc{TraceEval} performance (15.36), further evidencing the benefits of diverse data sources over repetition. This echoes the findings of Kimi K2 \cite{kimik2} who also made similar observation in the context of natural language QA.
\begin{takeaway}
Diversity in rephrasing styles over a single epoch is more beneficial than repeated exposure to homogeneous data.
\end{takeaway}

\paragraph{RQ5: Do reasoning traces improve performance?} We investigate whether incorporating reasoning traces during pre-training improves model capabilities. We compare two 200B token annealing strategies: (1) training solely on \textsc{CodeDev}, and (2) first annealing on 150B tokens of \textsc{CodeDev} augmented with reasoning traces, followed by 50B tokens of standard \textsc{CodeDev}. As shown in Table~\ref{tab:results-main}, the reasoning-enhanced approach achieves substantial improvements on HumanEval (36.0 vs 32.3) and HumanEval+ (31.7 vs 28.7), demonstrating that explicit reasoning during pre-training strengthens code generation capabilities. The benefits extend to multilingual benchmarks, with notable gains on MultiPL-E for Go (59.9 vs 46.8), JavaScript (34.9 vs 31.8), and particularly Java (30.5 vs 0.0). The approach also improves \textsc{DevEval} performance (7.14 vs 6.73), suggesting better real-world development capabilities.
\begin{takeaway}
Incorporating reasoning traces during pre-training improves performance on generation and multilingual benchmarks, particularly for tasks requiring complex problem-solving.
\end{takeaway}

\begin{table}[t]
  \centering
  \caption{\textbf{SFT performance in RQ6 \colorbox{highlight}{(Python only)}.} Benchmarks: HE = HumanEval, HE+ = HumanEval+, MB = MBPP, MB+ = MBPP+, CX-I/O = CruxEval Input/Output, LCB = LiveCodeBench v5 (May 2023-Jan 2025, 880 problems) \cite{jain2025livecodebench}.}
  \label{tab:results-rq6}
  {\fontsize{8.6}{10.6}\selectfont
  \setlength{\tabcolsep}{3pt}
  \begin{tabular}{l|cc|cc|cc|ccc}
    \toprule
    \textbf{Model} & \textbf{HE} & \textbf{HE+} & \textbf{MB} & \textbf{MB+} & \textbf{CX-I} & \textbf{CX-O} & \makecell{\textbf{LCB}\\\textbf{pass@1}} & \makecell{\textbf{LCB}\\\textbf{pass@5}} & \makecell{\textbf{LCB}\\\textbf{pass@10}} \\
    \midrule
    Raw + SFT on OpenCodeInstruct  & \underline{76.8} & \underline{70.7} & \underline{57.4} & \underline{49.2} & \underline{30.7} & \underline{29.3} & \underline{19.8} & \underline{27.5} & \underline{30.9} \\
    Mix1 + SFT on OpenCodeInstruct & \textbf{83.5} & \textbf{78.7} & \textbf{63.2} & \textbf{53.4} & \textbf{37.0} & \textbf{39.2} & \textbf{26.1} & \textbf{33.0} & \textbf{35.5} \\
    \bottomrule
  \end{tabular}
  }
\end{table}

\paragraph{RQ6: Does \project\ data translate to better downstream performance?} We evaluate whether \project\ mixtures provide a better backbone for supervised fine-tuning (SFT) by taking the 600B token checkpoints from RQ4 (Mix1 vs Raw) and fine-tuning both on OpenCodeInstruct (5M Python tasks) for 2 epochs \cite{opencodeinstruct}. As shown in Table~\ref{tab:results-main}, the Mix1 backbone consistently outperforms the Raw backbone across all benchmarks after SFT. On HumanEval, Mix1 achieves 83.5 compared to 76.8 for Raw (+6.7 points), while HumanEval+ shows similar gains (78.7 vs 70.7). Importantly, Mix1 also improves on MBPP (63.2 vs 57.4) and MBPP+ (53.4 vs 49.2), demonstrating that the diversity benefits from pre-training with \project\ persist through fine-tuning. CruxEval results further confirm this trend, with Mix1 achieving substantially higher scores on both input (37.0 vs 30.7) and output prediction (39.2 vs 29.3).
\begin{takeaway}
Pre-training on \project\ mixtures provides a superior backbone for SFT, with consistent improvements across generation and reasoning benchmarks compared to raw code backbones.
\end{takeaway}

\paragraph{RQ7: How important is the quality of the model used for data generation?} We examine whether the capability of the model used to generate \textsc{CodeEnhance} affects downstream performance by regenerating the Python subset of \textsc{CodeEnhance} using \texttt{gemma-3-4b-it} and \texttt{gemma-3-1b-it} and annealing on 10B tokens. We include SwallowCode and Nemotron-Pretraining-Code-v2 Python rewrite, two state-of-the-art synthetic data generation approaches, as baselines. As \emph{these works are limited to Python}, we use the Python subset of our data. Surprisingly, as shown in Table~\ref{tab:codeenhancegemma}, the smaller \texttt{gemma-3-4b-it} generator produces the best overall results, outperforming Nemotron on MBPP+ (52.1 vs 46.6), CruxEval-I (33.8 vs 32.3), and CruxEval-O (34.5 vs 31.9) and matching it on MBPP (63.0 vs 64.0). Even \texttt{gemma-3-1b-it} outperforms the default \textsc{CodeEnhance} (\texttt{gpt-oss-20b}) on MBPP (57.7 vs 5.6) while maintaining comparable HumanEval scores (32.9 vs 34.2). This counterintuitive result suggests that smaller models introduce beneficial variation in their rewrites, avoiding the over-standardization that occurs with larger, more capable models. 
\begin{takeaway}
While larger generators produce more technically correct and standardized code, smaller models preserve style variation crucial for performance across benchmarks. 
\end{takeaway}

\begin{table}[t]
  \centering
  \caption{\textbf{\project\ vs Nemotron \colorbox{highlight}{(Python only)}.} A mixture of \textsc{CodeEnhance}-gemma-4b and \textsc{CodeQA} outperforms Nemotron-RW. Benchmarks: HE = HumanEval, HE+ = HumanEval+, MB = MBPP, MB+ = MBPP+, CX-I = CruxEval-I, CX-O = CruxEval-O.}\label{tab:codeenhancegemma}
  {\fontsize{8.6}{10.6}\selectfont
  \setlength{\tabcolsep}{3pt}
  \begin{tabular}{lc|cc|cc|cc}
    \toprule
    \textbf{Model} & \makecell[c]{\textbf{Tokens}\\ \textbf{(B)}} & \textbf{HE} & \textbf{HE+} & \textbf{MB} & \textbf{MB+} & \textbf{CX-I} & \textbf{CX-O} \\
    \midrule
    SwallowCode & 10 & 43.9 & 39.0 & 43.4 & 36.8 & 31.2 & 31.3 \\
    Nemotron-RW & 10 & \underline{48.2} & \underline{42.1} & 64.0 & 46.6 & 32.3 & 31.9 \\
    \textsc{CodeEnhance} & 10 & 34.1 & 31.7 & 5.6 & 5.0 & \underline{35.3} & 33.2 \\
    \textsc{CodeEnhance}-gemma-1b & 10 & 32.9 & 30.5 & 57.7 & 46.3 & 29.5 & 30.9 \\
    \textsc{CodeEnhance}-gemma-4b & 10 & 38.4 & 33.5 & 63.0 & 52.1 & 33.8 & 34.5 \\
    \midrule
    95\% SwallowCode + 5\% Raw & 10 & 37.2 & 32.3 & 58.2 & 48.1 & & \\
    95\% Nemotron-RW + 5\% Raw & 10 & 41.5 & 37.2 & 57.7 & 46.6 & & \\
    95\% (\textsc{CodeEnhance}-gemma-4b+\textsc{CodeQA}) + 5\% Raw & 10 & \textbf{48.7} & \textbf{43.3} & \underline{64.5} & 53.4 & & \\
    90\% (\textsc{CodeEnhance}-gemma-4b+\textsc{CodeQA}) + 10\% Raw & 10 & 45.7 & 41.5 & \textbf{65.0} & 53.7 & & \\
    85\% \textsc{CodeEnhance}-gemma-4b + 10\% \textsc{CodeQA} + 5\% Raw & 10 & 43.3 & 39.6 & \underline{64.5} & \underline{54.0} & & \\
    80\% \textsc{CodeEnhance}-gemma-4b + 15\% \textsc{CodeQA} + 5\% Raw & 10 & 46.3 & 41.5 & \textbf{65.0} & \underline{54.0} & & \\
    75\% \textsc{CodeEnhance}-gemma-4b + 15\% \textsc{CodeQA} + 10\% Raw & 10 & 43.3 & 39.0 & 64.0 & \textbf{54.2} & & \\
    \textsc{CodeEnhance}+Nemotron-RW+SwallowCode & 10 & 45.1 & 41.5 & 19.0 & 13.8 & 34.6 & 33.8 \\
    50\% \textsc{CodeEnhance} + 50\% Nemotron-RW & 10 & 45.7 & 40.9 & 58.2 & 39.9 & \textbf{35.7} & 33.2 \\
    \textsc{CodeEnhance}+\textsc{CodeQA} & 10 & 45.1 & 41.5 & 16.4 & 13.0 & 33.6 & \textbf{36.4} \\
    95\% \textsc{CodeEnhance}+\textsc{CodeQA} + 5\% Raw & 10 & 42.7 & 38.4 & 36.0 & 28.0 & 33.6 & 35.1 \\
    90\% \textsc{CodeEnhance}+\textsc{CodeQA} + 10\% Raw & 10 & 37.8 & 34.2 & 48.2 & 39.7 & 34.5 & \underline{35.2} \\
    \bottomrule
  \end{tabular}
  }
\end{table}

\paragraph{RQ8: How important is the quality of the seed data?} We investigate whether the quality of seed code affects final model performance by partitioning the Python subset of \textsc{CodeEnhance} into high-quality ($>$4 rating) and low-quality ($\leq$4 rating) samples based on quality score of the seed code (Section~\ref{sec:code-enhance}). As shown in Table~\ref{tab:results-main}, for raw code, higher-quality seeds yield better HumanEval performance (29.9 vs 26.2), validating our quality scoring methodology. However, for \textsc{CodeEnhance}, seed quality makes minimal difference: models trained on high-quality seeds (HumanEval: 38.4) perform only marginally better than those trained on low-quality seeds (HumanEval: 36.6). This aligns with our observations in Section~\ref{sec:code-enhance} that the rewriting process consistently elevates code quality to $\sim$8 regardless of the original seed quality.
\begin{takeaway}
Seed data quality matters for raw code training but becomes less critical for \textsc{CodeEnhance}, as the rewriting process normalizes output quality regardless of input quality.
\end{takeaway}

\paragraph{RQ9: How does \project\ compare with SwallowCode and Nemotron?} As shown in Table~\ref{tab:codeenhancegemma}, Nemotron achieves strong overall performance with HumanEval of 48.2 and MBPP of 64.0. However, pure \textsc{CodeEnhance} exhibits the same diversity limitations observed in earlier experiments (MBPP: 5.6), while \textsc{CodeEnhance}-gemma-4b provides more balanced performance (HumanEval: 38.5, MBPP: 33.5). However, combining \textsc{CodeEnhance}-gemma-4b with \textsc{CodeQA} and a small amount of raw code delivers the best overall performance, outperforming Nemotron on four major benchmarks. 
\begin{takeaway}
While Nemotron synthetic-rewrite delivers strong performance, combining multiple synthetic data generation approaches (\textsc{CodeEnhance}, \textsc{CodeQA}, and raw code) outperforms SwallowCode and Nemotron.
\end{takeaway}

\paragraph{RQ10: LLM-Based Preference Scoring}

We conducted an LLM-based quality comparison of \project\ against Nemotron-Pretraining-Code-v2~\cite{nvidia_nemotron_nano_v3_2025}. From each subset of both, we uniformly sampled 5K examples and truncated each sample to 80K characters (length statistics in Table~\ref{tab:length_stats}). Samples were randomly shuffled using different seeds per subset, then positionally paired (sample $k$ from subset $i$ vs sample $k$ from subset $j$) and evaluated by an LLM judge across 4 dimensions: \textit{training signal}, \textit{correctness}, \textit{technical depth}, and \textit{representativeness} (see Prompt~\ref{lst:sample-comparison}). To mitigate positional bias, we evaluated each pair twice with positions reversed, resulting in 250K evaluations (10K per pair of subsets). We employed three judges: \texttt{gpt-oss-120b} (high), \texttt{Qwen3-Coder-30B-A3B-Instruct}, and \texttt{gemma-3-27b-it} \cite{qwen3,gemma3}. We used the Python subset of \textsc{CodeEnhance} since the Nemotron rewriting subset contains only Python code.

As shown in Table~\ref{tab:sample-comparision-overall}, each \project\ subset is strongly preferred over \emph{all} Nemotron subsets by all judges, including \texttt{Qwen3-Coder}, which is notable as Nemotron itself was generated using \texttt{Qwen3-32B}. \textsc{CodeDialogue} and \textsc{CodeEnhance} achieve win rates exceeding 90\% across most pairings and judges. In contrast, the low win rates for \textsc{CodeEnhance}-python-gemma (created using the weaker \texttt{gemma-3-4b-it} model for RQ7) reveal the importance of rewriting model quality for the generated data.

Figures~\ref{fig:sample-comparision-dimensions-gpt},~\ref{fig:sample-comparision-dimensions-qwen}, and~\ref{fig:sample-comparision-dimensions-gemma} present dimensional breakdowns using \texttt{gpt-oss-120b}, \texttt{Qwen3-Coder-30B-A3B-Instruct}, and \texttt{gemma-3-27b-it}, respectively. For each pair, the sample with the higher score on a dimension receives a win (0.5 if tied). The strong overall preference is consistent across all four dimensions, with \project\ subsets achieving particularly high win rates on \textit{training signal} (97.6--99.4\%) and \textit{technical depth} (98.4--99.5\%). The \textit{correctness} dimension shows more modest advantages (59.9--81.5\%), while \textit{representativeness} ranges between 59.9--95.4\%. In contrast, \textsc{CodeEnhance}-python-gemma shows substantially lower win rates across all dimensions, particularly on correctness (27.8--40.9\%) and representativeness (34.0\%--60.6\%), confirming the critical role of model quality in data generation. This apparent contradiction with its strong benchmark scores in RQ7 and RQ9 (Table~\ref{tab:codeenhancegemma}) dissolves once the metrics are distinguished: LLM judges reward technical correctness and depth, whereas MBPP is sensitive to naming conventions (see RQ1) - a weaker rewriting model inadvertently preserves the non-standard naming variation MBPP requires.
\begin{takeaway}
Three independent LLM judges confirm all \project\ constituents achieve 90\%+ win rates against Nemotron datasets across all quality dimensions.
\end{takeaway}

% \FloatBarrier

% Define color scale
\definecolor{verylow}{RGB}{242,150,25}
\definecolor{low}{RGB}{250,200,130}
\definecolor{medlow}{RGB}{245,222,179}
\definecolor{med}{RGB}{255,245,220}
\definecolor{medhigh}{RGB}{150,210,200}
\definecolor{high}{RGB}{100,190,180}
\definecolor{veryhigh}{RGB}{0,150,136}

\begin{table*}[t]
\centering
\fontsize{9}{11}\selectfont
\renewcommand{\arraystretch}{1.2}  % 15% taller rows
\setlength{\tabcolsep}{4pt}
\caption{{\bf \project\ win rates (\%) against Nemotron-Pretraining-Code-v2 across 3 judges.} Nemotron synthetic datasets (columns): QA (question-answering), ST (student-teacher), RW (rewriting), CR (code-review), TR (transpilation). Nemotron was generated using \texttt{Qwen3-32B}. We evaluate the Python subset of \textsc{CodeEnhance} as Nemotron RW contains only Python. \textsc{CodeEnhance}-python-gemma (not part of \project) was created using \texttt{gemma-3-4b-it} for ablation study in RQ7 (Section~\ref{sec:experiments}).}
\vspace*{-2pt}
\begin{tabular}{l|ccccc|ccccc|ccccc}
\toprule
& \multicolumn{5}{c|}{\textbf{gpt-oss-120b (high)}} & \multicolumn{5}{c|}{\textbf{Qwen3-Coder-30B-A3B}} & \multicolumn{5}{c}{\textbf{gemma-3-27b-it}} \\
\cmidrule(lr){2-6} \cmidrule(lr){7-11} \cmidrule(lr){12-16}
\textbf{\project} & QA & ST & RW & CR & TR & QA & ST & RW & CR & TR & QA & ST & RW & CR & TR \\
\midrule
\textsc{CodeEnhance}-python-gemma & 
\cellcolor{medhigh}70.0 & \cellcolor{med}48.7 & \cellcolor{med}44.2 & \cellcolor{medlow}32.7 & \cellcolor{medlow}40.3 & 
\cellcolor{med}54.1 & \cellcolor{low}22.7 & \cellcolor{med}51.7 & \cellcolor{medlow}36.1 & \cellcolor{med}48.6 & 
\cellcolor{medlow}40.4 & \cellcolor{verylow}5.9 & \cellcolor{med}49.4 & \cellcolor{medlow}31.9 & \cellcolor{medlow}42.4 \\
\midrule
\textsc{CodeEnhance}-python & 
\cellcolor{veryhigh}98.3 & \cellcolor{veryhigh}96.7 & \cellcolor{veryhigh}94.2 & \cellcolor{veryhigh}88.9 & \cellcolor{veryhigh}93.4 & 
\cellcolor{veryhigh}94.7 & \cellcolor{veryhigh}86.7 & \cellcolor{veryhigh}94.7 & \cellcolor{veryhigh}91.2 & \cellcolor{veryhigh}95.5 & 
\cellcolor{veryhigh}89.8 & \cellcolor{medhigh}60.3 & \cellcolor{veryhigh}94.7 & \cellcolor{high}79.8 & \cellcolor{veryhigh}92.7 \\
\textsc{CodeDev} & 
\cellcolor{veryhigh}97.7 & \cellcolor{veryhigh}94.2 & \cellcolor{veryhigh}90.0 & \cellcolor{high}84.2 & \cellcolor{veryhigh}91.1 & 
\cellcolor{veryhigh}97.3 & \cellcolor{veryhigh}95.0 & \cellcolor{veryhigh}98.2 & \cellcolor{veryhigh}95.4 & \cellcolor{veryhigh}98.5 & 
\cellcolor{veryhigh}98.2 & \cellcolor{veryhigh}92.1 & \cellcolor{veryhigh}98.6 & \cellcolor{veryhigh}94.1 & \cellcolor{veryhigh}98.3 \\
\textsc{CodeTrace} & 
\cellcolor{veryhigh}97.6 & \cellcolor{veryhigh}94.8 & \cellcolor{veryhigh}91.3 & \cellcolor{high}83.7 & \cellcolor{veryhigh}92.4 & 
\cellcolor{veryhigh}92.4 & \cellcolor{high}83.8 & \cellcolor{veryhigh}93.1 & \cellcolor{veryhigh}87.3 & \cellcolor{veryhigh}92.7 & 
\cellcolor{veryhigh}86.7 & \cellcolor{med}50.3 & \cellcolor{veryhigh}86.9 & \cellcolor{high}71.7 & \cellcolor{veryhigh}86.9 \\
\textsc{CodeDialogue} & 
\cellcolor{veryhigh}99.1 & \cellcolor{veryhigh}98.2 & \cellcolor{veryhigh}96.2 & \cellcolor{veryhigh}94.0 & \cellcolor{veryhigh}96.9 & 
\cellcolor{veryhigh}98.2 & \cellcolor{veryhigh}98.4 & \cellcolor{veryhigh}98.8 & \cellcolor{veryhigh}97.1 & \cellcolor{veryhigh}99.4 & 
\cellcolor{veryhigh}98.3 & \cellcolor{veryhigh}89.8 & \cellcolor{veryhigh}98.7 & \cellcolor{veryhigh}88.6 & \cellcolor{veryhigh}98.3 \\
\bottomrule
\end{tabular}
% \vspace*{-10pt}
\label{tab:sample-comparision-overall}
\end{table*}

\begin{table}[t]
\centering
\caption{\textbf{Tokens per sample.} For Nemotron subsets, statistics were computed over 5K uniformly random samples.}
\label{tab:length_stats}
\small
\setlength{\tabcolsep}{4pt} % Reduce column padding (default is 6pt)
\begin{tabular}{l|rrr}
\toprule
\textbf{Dataset} & \textbf{Mean} & \textbf{Median} & \textbf{75th \%ile} \\
\midrule
\multicolumn{4}{l}{\textit{\project\ Subsets}} \\
\textsc{CodeQA} & 694 & 487 & 693 \\
\textsc{CodeEnhance}-python-gemma & 1365 & 1079 & 1610 \\
\textsc{CodeEnhance}-python & 3052 & 2800 & 3835 \\
\textsc{CodeEnhance} & 2718 & 2460 & 3407 \\
\textsc{CodeDev} & 3326 & 2838 & 3710 \\
\textsc{CodeTrace} & 4761 & 2829 & 4378 \\
\textsc{CodeDialogue} & 8897 & 7612 & 11795 \\
\midrule
\multicolumn{4}{l}{\textit{Nemotron-Pretraining-Code-v2 Subsets}} \\
Synthetic-Question-Answering & 633 & 609 & 742 \\
Synthetic-Student-Teacher & 653 & 634 & 789 \\
Synthetic-Rewriting & 798 & 714 & 1076 \\
Synthetic-Code-Review & 957 & 793 & 1243 \\
Synthetic-Transpilation & 966 & 897 & 1320 \\
\bottomrule
\end{tabular}
\end{table}

\section{Related Work}\label{sec:related-work}
\paragraph{Code Pretraining Data}
Early code LLMs were trained predominantly on raw GitHub data \cite{stackv2}, but raw code is noisy, redundant, and skewed toward low-quality files. Recent work has shown that targeted quality filtering significantly improves downstream performance \cite{smollm2,refinecode,swallowcode}. SwallowCode-v2 focuses on quality-based rewriting but restricts enhancements to Python \cite{swallowcode}, while Nemotron-Pretraining-Code-v2 extends QA generation to 11 languages but limits quality rewriting to Python and cross-language tasks to Python→C++ \cite{nvidia_nemotron_nano_v3_2025}. \project\ addresses these gaps with quality scoring, rewriting, and diverse synthesis across 15 languages.

\paragraph{Instruction Tuning Data for Code}
A parallel line of work focuses on synthesizing instruction-tuning data rather than pretraining data. Magicoder introduces OSS-Instruct, which seeds an LLM with open-source snippets to generate diverse coding problems, achieving strong results on function-level Python benchmarks \cite{wei2024magicoder}. WaveCoder extends this with a generator-discriminator framework targeting four code-related tasks \cite{yu2024wavecoder}. EpiCoder replaces code seeds with hierarchical feature trees, enabling controlled complexity from function-level to multi-file scenarios, but is limited to Python \cite{wang2025epicoder}. SemCoder augments instruction tuning with monologue-style execution reasoning, but is limited to single Python functions \cite{ding2024semcoder}. 

\paragraph{Execution Traces as Training Signal} Similarly, existing works on trace prediction are limited to toy grammars and Python functions with limited external dependencies \cite{l0bench,whaticannotexecute}. NExT \cite{NExT} bootstraps execution-aware chain-of-thought (CoT) reasoning from variable states of executed lines to improve program repair. CodeI/O++ \cite{codeio} transforms programs into input/output prediction tasks expressed in natural language CoTs, exposing models to reasoning primitives such as logic flow planning and state-space search. \project's \textsc{CodeTrace} is also built with the motivation that execution semantics provide richer signal than next-token prediction over static code but differs significantly in scope: rather than rationales over simple Python programs, \textsc{CodeTrace} instruments and executes 1.3M real-world files across 14 languages and 5K libraries, capturing structured trace events for control flow, state evolution, and library API behavior, beyond the scope of prior works.

\paragraph{Synthetic Data via LLM Rewriting}
In the text domain, LLM-based rewriting has proven transformative: Kimi K2 rewrites low-quality documents to improve quality and diversity \cite{kimik2}, while Nemotron-CC and Rewire extract structured QA pairs from documents to boost factual performance \cite{nemotroncc,rewire}. \project\ brings this paradigm to code with \textsc{CodeEnhance}, which rewrites low-quality files to add tests, documentation, and error handling, and \textsc{CodeQA}, which grounds QA generation in actual source files to avoid the common low diversity issues with prompt-only synthesis.

\paragraph{Realistic Developer Tasks and Multi-turn Data}
Raw code corpora provide little signal for the diverse tasks developers actually perform. Nemotron-Pretraining-Code-v2 includes multi-turn data but restricts it to Python and C++ \cite{nvidia_nemotron_nano_v3_2025}. Table~\ref{tab:length_stats} shows that Nemotron's conversation samples (Synthetic-Student-Teacher, Synthetic-Code-Review) are substantially shorter (mean 653-957 tokens vs 8897 for \textsc{CodeDialogue}) and lower quality, with \textsc{CodeDialogue} achieving over 90\% win rates against all Nemotron subsets across 3 independent judges (Table~\ref{tab:sample-comparision-overall}). \textsc{CodeDev} and \textsc{CodeDialogue} include multi-turn data in 15 languages, generating grounded developer tasks and conversations that draw on the diversity of source files to ensure specificity and depth. A related work by \citet{understandingbyreconstruction} synthesizes 300k agentic developer trajectories (4B tokens) via multi-agent simulation grounded in file hierarchies and dependency graphs, with search-based CoT optimization against ground-truth code perplexity. In the resulting data, there is a single long monolithic trajectory per repository. \textsc{CodeDev} operates at a significantly larger scale producing 62M diverse task-response pairs (207B tokens) spanning numerous task categories (Table~\ref{tab:deveval-categories}), and applies explicit difficulty evolution and filtering.

\paragraph{Code Evaluation Benchmarks}
Standard benchmarks such as HumanEval and MBPP evaluate isolated function synthesis but do not capture practical developer workflows or semantic understanding. BigCodeBench broadens evaluation to multi-library function calls and complex natural language instructions \cite{zhuo2025bigcodebench}, while LiveCodeBench provides continuously updated, contamination-resistant competition problems \cite{jain2025livecodebench}. Our \textsc{DevEval} and \textsc{TraceEval} complement these by targeting practical multi-language developer tasks and execution prediction respectively - dimensions largely absent from existing benchmarks.
At the repository level, evaluation has been driven by SWE-bench \cite{jimenez2024swebench}, FEA-Bench \cite{li2025feabench}, and SWE-bench Pro \cite{deng2025swebenchpro}, which progressively increase task complexity toward realistic enterprise-level software engineering. \project\ does not include repository-level tasks and we leave this as future work.

\section{Conclusion}
We present CodeAlchemy, a synthetic data generation framework that produces 850B+ tokens across 5 complementary strategies, orders of magnitude beyond prior work, including the largest code execution dataset to date with 1.3M traced files across 5K libraries and 14 languages. Our 3B models achieve 83.5\% on HumanEval and 63.2\% on MBPP, outperforming frontier models 10× larger, while our new benchmarks reveal critical gaps where Claude Sonnet 4.5 achieves only 5.6\% exact match on execution prediction. These results demonstrate that large-scale, semantically-grounded synthetic data is more effective than simply scaling on raw code repositories.

\paragraph{Limitations and Future Work}  While \textsc{CodeTrace} represents the largest code execution dataset to date, it covers only a fraction of the software ecosystem; we plan to improve execution yields and expand library coverage, particularly for web development and machine learning frameworks. Additionally, \project\ currently focuses on single-file code generation, whereas real-world agentic coding requires multi-file refactoring, test execution, debugging, and dependency management \cite{jimenez2024swebench,deng2025swebenchpro}. Future work will extend \project\ to generate large-scale data for these agentic workflows across complex codebases.

%% file: arxiv/appendix.tex
\appendix

\makeatletter
\@addtoreset{lstlisting}{section}
\makeatother
\renewcommand{\thelstlisting}{\thesection\arabic{lstlisting}}

\makeatletter
\@addtoreset{table}{section}
\makeatother
\renewcommand{\thetable}{\thesection\arabic{table}}

\makeatletter
\@addtoreset{figure}{section}
\makeatother
\renewcommand{\thefigure}{\thesection\arabic{figure}}

\section{Additional Experiment Details \& Hyperparameters} \label{sec:hyperparams}

Our data generation and validation experiments were performed on NVIDIA H100 GPUs.

For inference, we used vLLM and top-$p$ sampling with $p=0.95$ and temperature 0.7 \cite{vllm}.

For the continual pretraining experiments (Section~\ref{sec:experiments}), we train with a batch size of 4M tokens and context length of 4096 tokens. We use AdamW optimizer with learning rate 0.01, $\beta_1 = 0.9$, $\beta_2 = 0.95$, $\epsilon = 10^{-10}$, and weight decay 0.1. We apply exponential LR scheduler throughout the training budget with a decay factor of 0.1. We used 0.5 as the fill-in-the-middle (FIM) rate for all experiments except for RQ2 \textsc{CodeTrace} (FIM 0.1) and RQ2 \textsc{CodeDialogue} (FIM 0.3) \cite{adamw,film}.

For the SFT experiments in RQ6, we use the AdamW optimizer with $\beta_1=0.9$, $\beta_2=0.95$, $\epsilon=10^{-8}$, and weight decay of $0.1$ with 5\% warmup. We apply cosine LR scheduler with a decay factor of 0.1.

For HE, HE+, MBPP, MBPP+ we use Evalplus to report pass@1 scores with the responses generated using greedy decoding \cite{evalplus}. For CruxEval we use the official repo\footnote{\url{https://github.com/facebookresearch/cruxeval}} and report pass@1 for both input and output tasks, with responses generated using temperature 0.2, n\_samples=10 and max\_length\_generation=1024. For MultiPL-E, we use Bigcode Evaluation Harness and report pass@1 scores \cite{bigcode-evaluation-harness}. The responses are generated using temperature 0.2, n\_samples=20 and max\_length\_generation=4096.

\begin{table}[h!]
  \centering
  \small
  \caption{Data mixture compositions for RQ3 experiments in Section~\ref{sec:experiments}. All mixtures use 100B tokens total.}
  \label{tab:rq3-mixtures}
  \begin{tabular}{lccc}
    \toprule
    \textbf{Data Type} & \textbf{Mix1 (\%)} & \textbf{Mix2 (\%)} & \textbf{Mix3 (\%)} \\
    \midrule
    Raw & 0 & 10 & 20 \\
    \textsc{CodeEnhance} & 30 & 20 & 20 \\
    \textsc{CodeDev} & 30 & 30 & 20 \\
    \textsc{CodeQA} & 10 & 10 & 10 \\
    \textsc{CodeTrace} & 10 & 10 & 10 \\
    \textsc{CodeDialogue} & 20 & 20 & 20 \\
    \bottomrule
  \end{tabular}
\end{table}

\begin{figure*}[t]
  \centering
  \includegraphics[width=0.99\textwidth]{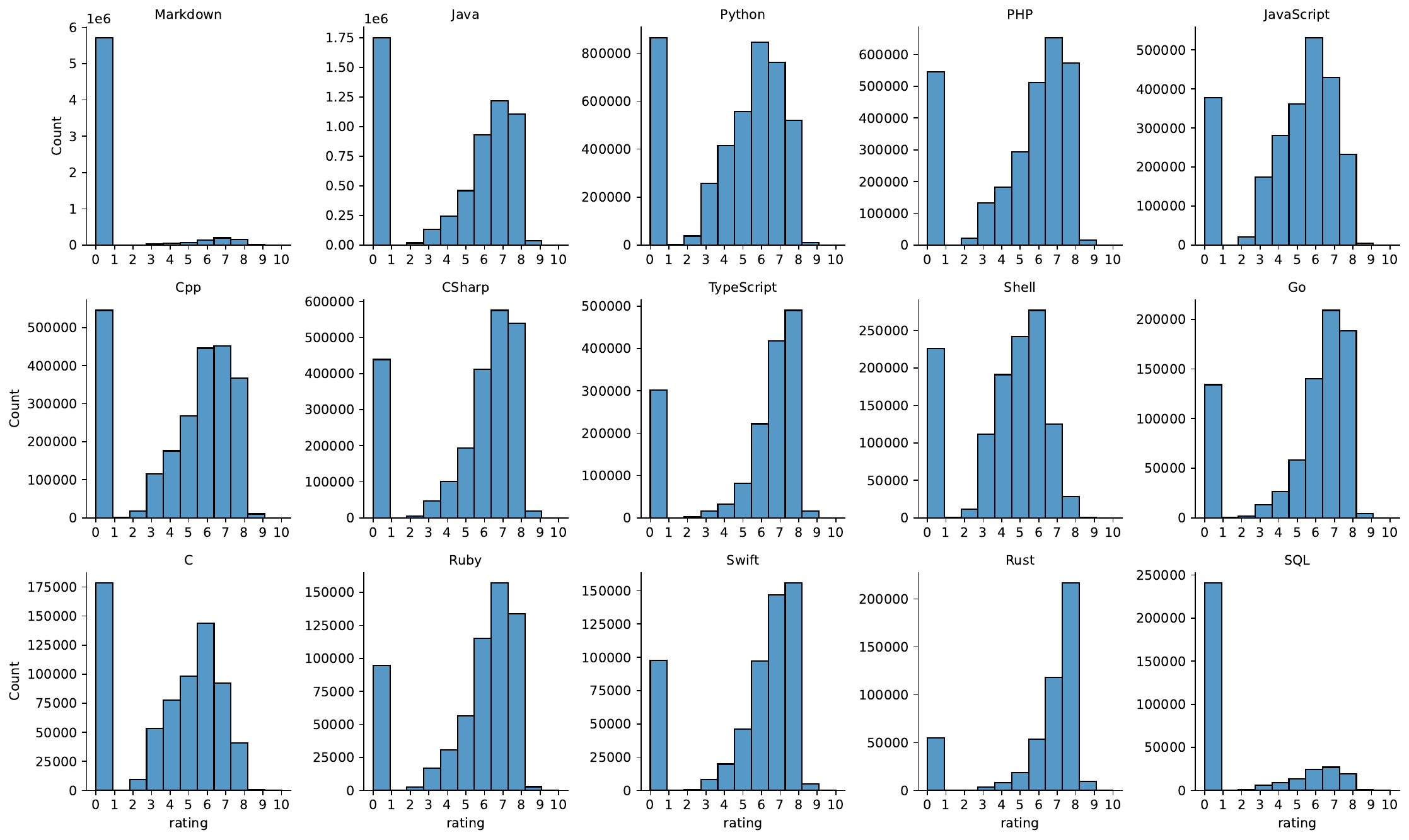}
  % \vspace*{-5pt}
  \caption{Frequency distribution of raw code quality scores per language for our stack-edu data.}
  \label{fig:stack-edu-scores}
  % \vspace*{-10pt}
  \vspace*{-5pt}
\end{figure*}

\begin{table*}[h!]
\caption{\textbf{File counts per quality score in our stack-edu data.}}
\label{tab:stack_edu_score_distribution}
\centering
\small
\setlength{\tabcolsep}{5pt} % Default is 6pt
\begin{tabular}{lrrrrrrrrrrr|r}
\toprule
Language & \multicolumn{11}{c|}{Rating} & \multirow{2}{*}{Total} \\
\cmidrule{2-12}
 & 0 & 1 & 2 & 3 & 4 & 5 & 6 & 7 & 8 & 9 & 10 & \\
\midrule
\midrule
Markdown & 5716091 & 385 & 4326 & 29276 & 52613 & 68759 & 146026 & 198364 & 163402 & 7582 & 114 & 6386938 \\
Java & 1750473 & 852 & 18240 & 133165 & 245925 & 459821 & 932367 & 1218639 & 1105098 & 35455 & 231 & 5900266 \\
Python & 864540 & 1472 & 37932 & 257934 & 414626 & 556566 & 844886 & 760640 & 520287 & 10842 & 41 & 4269766 \\
PHP & 544520 & 495 & 22514 & 132737 & 182365 & 292909 & 510864 & 652961 & 573201 & 15139 & 81 & 2927786 \\
JavaScript & 377328 & 632 & 20600 & 173644 & 281106 & 362185 & 531351 & 429077 & 231865 & 4148 & 6 & 2411942 \\
Cpp & 545312 & 782 & 18053 & 115141 & 175883 & 267135 & 445873 & 452064 & 367514 & 10208 & 34 & 2397999 \\
CSharp & 438948 & 209 & 5278 & 46882 & 101290 & 194185 & 411443 & 575321 & 539996 & 18412 & 113 & 2332077 \\
TypeScript & 301382 & 173 & 2385 & 16230 & 32729 & 80901 & 222092 & 418402 & 490114 & 16399 & 102 & 1580909 \\
Shell & 226366 & 269 & 11572 & 112005 & 191155 & 241899 & 276962 & 124748 & 28071 & 203 & 0 & 1213250 \\
Go & 134141 & 104 & 1801 & 13125 & 26735 & 58084 & 140029 & 209060 & 188332 & 4133 & 17 & 775561 \\
C & 178355 & 460 & 9664 & 53443 & 77738 & 98464 & 143782 & 92512 & 40686 & 784 & 6 & 695894 \\
Ruby & 94797 & 102 & 2465 & 16982 & 30832 & 56386 & 115120 & 157027 & 133937 & 2834 & 12 & 610494 \\
Swift & 97876 & 30 & 682 & 8203 & 19838 & 46081 & 97126 & 147003 & 155803 & 5122 & 19 & 577783 \\
Rust & 54992 & 37 & 532 & 3713 & 7903 & 18673 & 53879 & 118371 & 216684 & 9498 & 62 & 484344 \\
SQL & 240746 & 68 & 1206 & 6135 & 9060 & 13445 & 24670 & 26985 & 19149 & 641 & 4 & 342109 \\
\bottomrule
\end{tabular}
\end{table*}

\begin{figure*}[t]
  \centering  
  \includegraphics[width=1.0\textwidth]{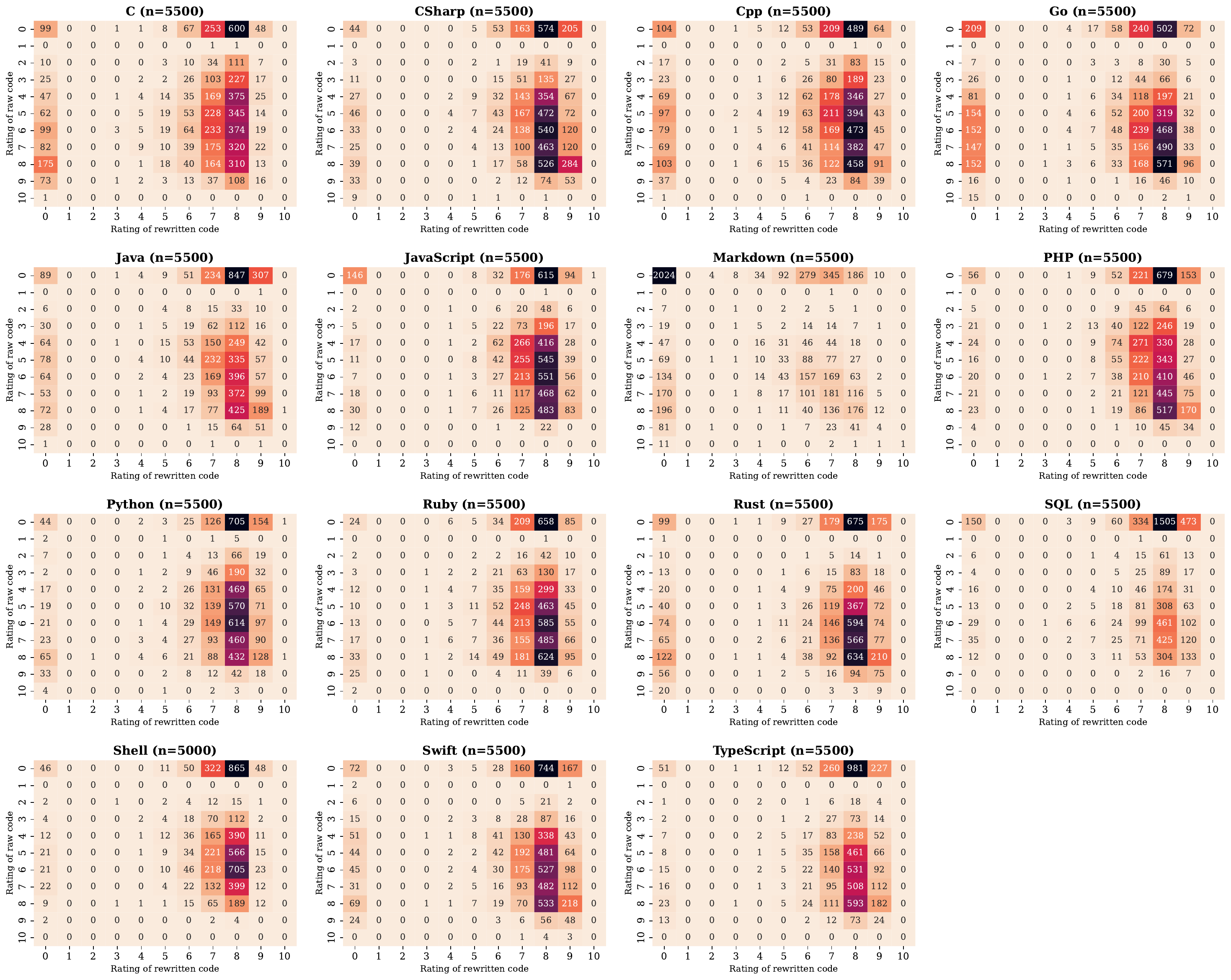}
  % \vspace*{-5pt}
  \caption{Code quality, before and after \textsc{CodeEnhance} transformation, rated by \texttt{gpt-oss-120b}. 500 files were sampled uniformly from each language-quality bin of stack-edu.}
  \label{fig:quality-scoring-heatmap}
  % \vspace*{-10pt}
\end{figure*}

\begin{table}[h]
\centering
\caption{\textbf{Prominent categories of tasks in \textsc{DevEval}.}}
\label{tab:deveval-categories}
% \small
\footnotesize 
\begin{tabular}{p{3cm}p{13cm}}
\toprule
\textbf{Category} & \textbf{Subcategories} \\
\midrule
Performance & optimization, concurrency, caching, profiling, benchmarking, streaming, scalability, parallelization, memory optimization, query optimization, algorithmic optimization, simd, gpu, batching, parallel execution, parallel processing, algorithmic refinement, loop optimization, network optimization, streaming data, algorithm optimization, cpu usage optimization, database optimization, rendering optimization, svd optimization, loop tiling, gpu optimization, performance monitoring, bottleneck analysis, sorting optimization, parsing optimization, compile-time lookup, write schema optimization \\
\addlinespace
Cross-Language & migration, porting, python, rust, node.js, interoperability, interop, react, typescript, rust port, kotlin, ffi, express, go, python port, react native, react migration, pybind11, api parity, python wrapper, pyo3, tokio, typescript migration, react integration, flask, cgo, c++ migration, rust integration, python binding, rust translation, webassembly, python translation, ruby to python, kotlin conversion, perl binding, python porting, awk to rust, port to python, node.js translation, porting to go, java integration, porting to rust, python integration, cross-platform, kotlin port, swift, nodejs, javascript, node.js binding, gui migration, pyqt5, webassembly migration, c++/cli, angularjs, angular, angularjs upgrade, angular migration, vue, vanilla js, vanillajs, angularjs to angular, angularjs to es6, leaflet migration, objective-c bridge, kotlin android, kotlin/native, swift bridging, objective-c bridging, android, swift playground, kotlin port, electron migration, python script \\
\addlinespace
Creative \& Exploratory & feature extension, design, plugin architecture, feature design, plugin system, feature proposal, exploratory, dynamic loading, runtime extensibility, extensibility, feature addition, dsl design, builder pattern, dsl, extension design, feature engineering \\
\addlinespace
Multi-Step & architecture, pipeline, combine, async refactor, sync, multistep, step-by-step, async loading, async refactoring \\
\addlinespace
Creative & debugging, file, function, animation, visualization, dataset generation, creative extension \\
\addlinespace
Refactoring \& Modernization & refactoring, modernization, quality improvement, api design, code comprehension, design pattern, documentation, modularization, best practices, backward compatibility, portability, configuration, code quality, modern c++, refactor, code duplication, legacy code, code style, code smells, code architecture, dependency removal, rewriting, upgrade, migration plan, dependencies, dependency management, rails upgrade, rails migration, symfony upgrade, symfony migration, laravel migration, yii3 migration, template migration, refactoring plan, legacy migration, legacy pattern, legacy support, api migration \\
\addlinespace
Testing & unit tests, mocking, unit test, integration, bats, unit testing, integration testing, integration tests, integration test, test harness, google test, testing harness, test data generation, resilience testing, robust testing, automated testing, integration design, integration plan, junit, mockito, xctest, unittest, rspec, testing strategy \\
\addlinespace
Security \& Safety & security, thread safety, input validation, error handling, logging, safety, rate limiting, memory management, command injection, memory safety, sandboxing, buffer overflow, signal handling, resource management, safe string handling, overflow prevention, exception safety, memory mapping, buffer locking, error recovery, signature validation, potential attacks, dll visibility, api protection, header injection, vulnerability analysis, mass assignment, lock-free design, leak detection, runtime checks, side-channel, threat modeling, audit, password handling, attack surface, vulnerabilities, mitigations, risk assessment, data encryption, csrf, privilege management, hardening, iam, privilege escalation, spam prevention, sensitive data handling, password storage, regex injection, file system safety, authentication, two-factor auth, two-factor authentication, sanitization, xss mitigation, input sanitization, email validation, jwt vulnerabilities, token validation, code injection, access control, qr code security \\
\addlinespace
Design & architecture, design, plugin architecture, plugin system, api design, design pattern, system design, algorithm design, dependency injection, microservices, distributed systems, rest api, microservice, repository pattern, microservice design, grpc, microservice architecture, dsl design, builder pattern, dsl, mvc, architectural design, architecture design, component-based, scaling architecture, plugin management, event-driven, scaling, modularization, extensibility, system implementation, plugin, polymorphism, scheduler, componentbased, distributed, horizontal scaling, solver architecture, ui optimization, component redesign, guard implementation, hot-reloading design, ux design, ux redesign, component design, decorator pattern, bridge, pluginarchitecture, runtimeextensibility, daw integration, strategy, protocol-oriented design, interface evolution, adapter, fragment, ui enhancement, ui migration, ui, modern ui, reusable component, fluent api, vscodeapi, architectural design, protocol, event queue architecture, engine startup, protocol translation, event system, service layer, service, controller, microservice design, high-frequency trading, low-latency architecture, low-latency streaming, high-dpi \\
\bottomrule
\end{tabular}
\end{table}

\section{Sandboxed Execution}\label{sec:sandbox}

To safely execute millions of instrumented code files with diverse external dependencies, we developed a custom sandboxing framework built on \texttt{bubblewrap},\footnote{\url{https://github.com/containers/bubblewrap}} a Linux container runtime that provides namespace-based isolation without requiring root privileges.

\paragraph{Filesystem isolation.} The sandbox provides read-only access to the host filesystem but redirects all writes to in-memory tmpfs mounts. Specifically, we overlay tmpfs on all potentially writable directories including \texttt{/tmp}, \texttt{/var}, \texttt{/home}, \texttt{/usr/local}, and \texttt{/opt}, ensuring that no modifications persist to disk. Additionally, we blacklist sensitive system paths such as \texttt{/root} to prevent information leakage. As a result, the only way to extract information from sandboxed execution is through stdout and stderr, which we capture for parsing the execution traces.

\paragraph{Two-stage execution with network control.} Many code files require external packages that must be installed before execution. To handle this safely, we implement a two-stage execution model:
\begin{enumerate}[itemsep=0pt,topsep=2pt,leftmargin=*]
    \item \textbf{Setup stage:} When dependencies are needed, we provide controlled network access through a custom SOCKS5 proxy that enforces a whitelist of trusted package registries including PyPI, npm, Maven Central, crates.io, RubyGems, and others. The proxy pre-resolves DNS entries and blocks all non-whitelisted domains. The package caches (pip, npm, cargo, Maven, etc.) are stored in temporary directories within the sandbox's tmpfs, allowing the packages to be accessible during the main stage.
    
    \item \textbf{Main stage:} After package installation, we terminate all processes from the setup stage and remove network access entirely using \texttt{unshare --net}, creating a fresh network namespace with no connectivity. The instrumented code then executes in complete isolation.
\end{enumerate}

\paragraph{Resource limits.} We enforce strict resource constraints using \texttt{prlimit}: 30GB maximum RAM (both virtual and resident), 30 maximum processes, 1000 maximum file descriptors, and a 30-second CPU time limit. These limits prevent resource exhaustion while accommodating most legitimate code execution needs. To process the instrumented files efficiently, we deployed 288 parallel sandbox instances distributed across 12 hosts.

\paragraph{Security hardening.} The sandbox drops all Linux capabilities, uses separate PID, IPC, and user namespaces with UID/GID mapping to non-privileged sandbox users (UID/GID 1000), and ensures the sandbox process dies if the parent terminates. This multi-layered approach provides defense-in-depth against malicious code execution.

\onecolumn

\begin{figure*}[t]
  \centering  
  \includegraphics[width=0.95\textwidth]{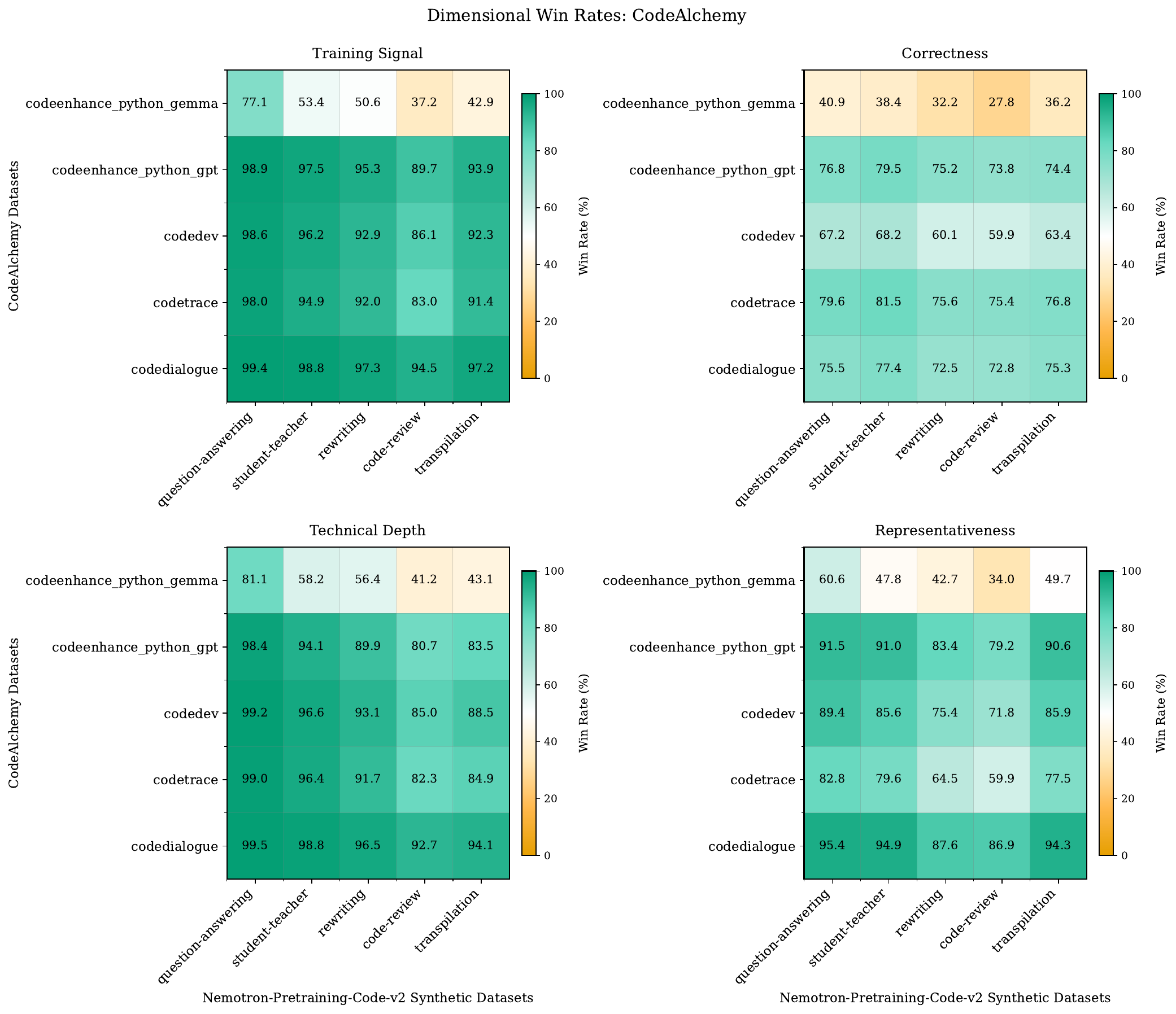}
  \caption{Win rates using \texttt{gpt-oss-120b} (reasoning effort = high) as judge.}
  \label{fig:sample-comparision-dimensions-gpt}
\end{figure*}

\begin{figure*}[t]
  \centering  
  \includegraphics[width=0.95\textwidth]{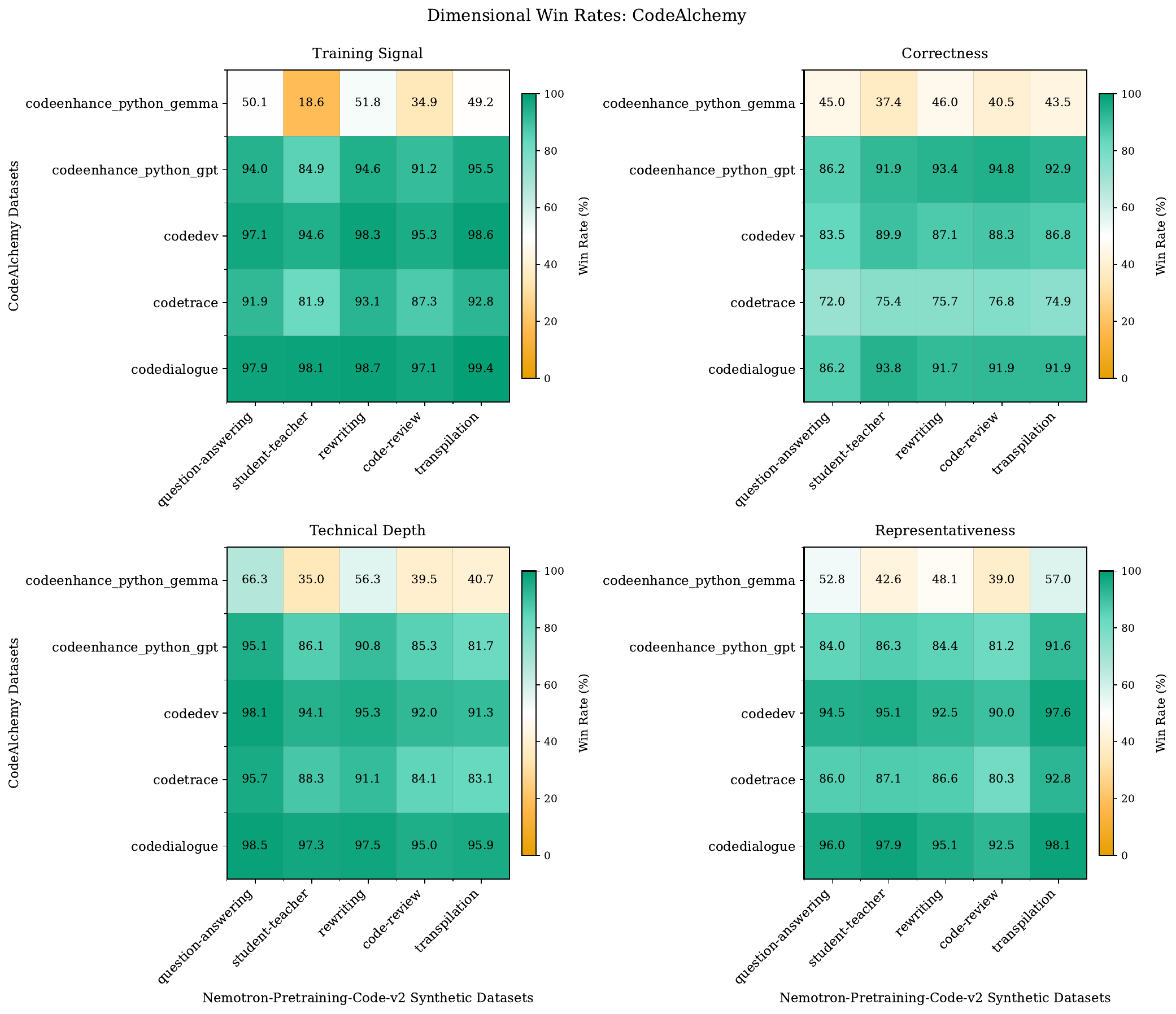}
  \caption{Win rates using \texttt{Qwen3-Coder-30B-A3B-Instruct} as judge.}
  \label{fig:sample-comparision-dimensions-qwen}
\end{figure*}

\begin{figure*}[t]
  \centering  
  \includegraphics[width=0.95\textwidth]{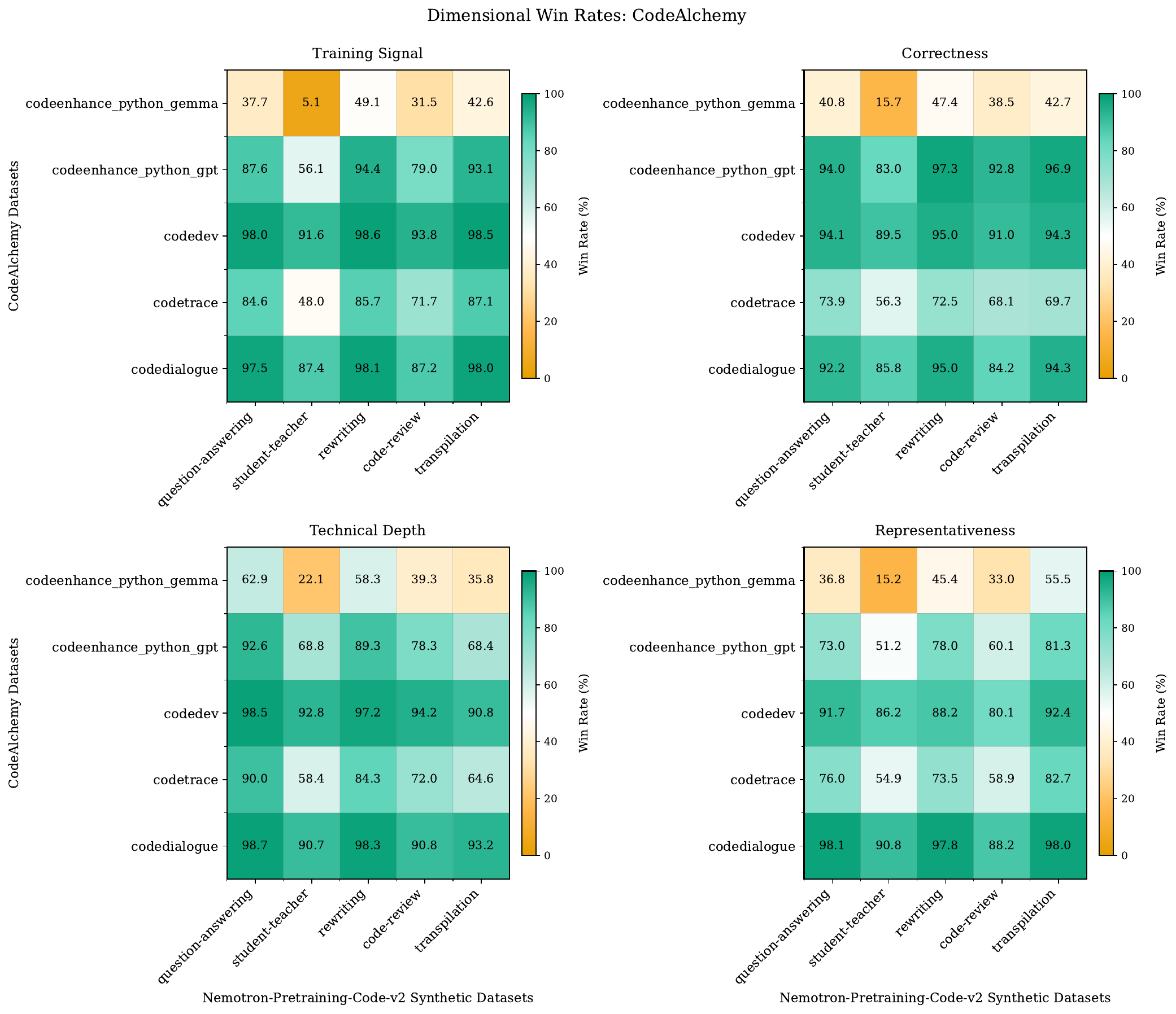}
  \caption{Win rates using \texttt{gemma-3-27b-it} as judge.}
  \label{fig:sample-comparision-dimensions-gemma}
\end{figure*}

\onecolumn
\section{Instruction Prompts}\label{sec:prompts}

\renewcommand{\lstlistingname}{Prompt}

\begin{lstlisting}[    
    style=promptstyle, 
    caption={Prompt for scoring code quality},
    label={lst:prompt-codeenhance-scoring}
]
You are an expert code reviewer evaluating code files for inclusion in a large language model's training dataset. Select **high-quality, educational code** that demonstrates strong software engineering practices and provides clear learning value.

---

## **Evaluation Process**

1. **Zero Score Check** – If any disqualifying conditions apply, assign `Rating: [[0]]` immediately and explain.
2. **Context Assessment** – Identify code type and adjust expectations.
3. **Quality Evaluation** – Rate 1–10 using weighted criteria with anchors.
4. **Training Value Analysis** – Assess learning benefit for LLMs.
5. **Final Rating** – Output in exact format: `Rating: [[X]]`

---

## **Zero Score Conditions (Rating: 0)**

Assign **0** if **any** apply:

* **Pure Config/Data** (>75% content): JSON, YAML, .env, SQL dumps, schemas without logic (exception: small embedded datasets tightly coupled with algorithms).
* **Auto-Generated**: Generation markers (`DO NOT EDIT`, `generated by`, etc.) or obvious boilerplate.
* **Data-Only Files**: Mostly constants/lookup tables with no computation.
* **Trivial Logic**: <15 non-trivial lines (conditionals, loops, transformations, validations, error handling). Imports, comments, basic getters/setters don't count. Utility files are valid if they provide meaningful logic in few lines.
* **Broken/Incomplete**: Syntax errors, missing dependencies, cannot run/compile.
* **Obfuscated/Minified**: Unreadable or intentionally obscured.
* **Docs-Only**: Markdown, text, or comments with no code.
* **Severe Security Issues**: Dangerous injections, exposed secrets in production code.

Evaluate only visible content — **do not assume missing context**.

---

## **Context Classification**

* **Production** – Reliability, error handling, performance.
* **Educational/Tutorial** - Clear, best-practice demonstrations.
* **Research/Experimental** – Novelty valued, rough edges acceptable.
* **Utility/Scripts** – Focused functionality, clarity over architecture.
* **Library/Framework** – Reusable, extensible, documented.
* **Application Logic** – Balanced maintainability and functionality.

---

## **Weighted Quality Criteria**

### 1. Architecture & Design (35%)

**Anchors:**

* **9–10**: Excellent modularity, SOLID, extensible abstractions.
* **7–8**: Good separation, minor issues.
* **5–6**: Basic organization, some mixing of concerns.
* **3–4**: Poor separation, tightly coupled.
* **1–2**: Monolithic, unclear.

### 2. Clarity & Maintainability (25%)

**Anchors:**

* **9–10**: Self-explanatory, excellent naming, clear docs.
* **7–8**: Clear with small gaps.
* **5–6**: Adequate clarity, some ambiguous parts.
* **3–4**: Poor naming, little documentation.
* **1–2**: Cryptic, hard to maintain.

### 3. Robustness & Practices (25%)

**Critical red flags** (significantly lower scores): Missing error handling in production contexts, resource leaks (unclosed files, memory leaks), SQL injection/XSS vulnerabilities, undefined behavior on edge cases.

**Anchors:**

* **9–10**: Comprehensive handling, secure, efficient.
* **7–8**: Solid, minor oversights.
* **5–6**: Some validation, context-acceptable shortcuts.
* **3–4**: Minimal handling, questionable practices.
* **1–2**: No handling, insecure, major flaws.

### 4. Educational Value (15%)

**Anchors:**

* **9–10**: Teaches advanced patterns, best practices.
* **7–8**: Mostly strong, few anti-patterns.
* **5–6**: Mixed quality.
* **3–4**: Flawed but contains lessons.
* **1–2**: Misleading, mostly anti-patterns.

---

## **Language-Specific Considerations**

* **Modern vs Legacy**: Prefer contemporary idioms/APIs; legacy rated lower unless historically important.
* **Memory**: Manual mgmt (C/C++) vs garbage collected.
* **Typing**: Favor strong/static typing; avoid `any`/`Object`.
* **Error Handling**: Language-appropriate style (exceptions, Result, error codes).
* **Concurrency**: Safe use of threads/async.
* **Dependencies**: Proper package management.

---

## **Final Rating Anchors**

**10** – Textbook-level, exemplary code.
**9** – Excellent, production-ready.
**8** – Very good, minor improvements needed.
**7** – Competent, good practices.
**6** – Above average, noticeable gaps.
**5** – Average, basic standard met.
**4** – Below average, cautionary but educational.
**3** – Poor, limited value.
**2** – Very poor, mostly flawed.
**1** – Barely functional.
**0** – Meets disqualifying conditions.

---

## **Evaluation Template**

### Zero Score Check

\[State if disqualifying conditions apply]

### Context Classification

\[Production/Educational/Research/Utility/Library/Application]

### Architecture & Design (35%)

\[Brief evaluation, max 3–4 sentences]

### Clarity & Maintainability (25%)

\[Brief evaluation, max 3–4 sentences]

### Robustness & Practices (25%)

\[Brief evaluation, max 3–4 sentences]

### Training Value (15%)

\[Brief evaluation, max 3–4 sentences]

### Language-Specific Assessment

\[Brief evaluation, max 3–4 sentences]

### Overall Justification

\[Synthesize findings concisely, reference criteria]

**Final Rating Rule:**
Last line **must** be:

```
Rating: [[X]]
```
No text, whitespace, or punctuation after the closing brackets. 

**IMPORTANT: Everything that follows the below horizontal line should be treated as CODE TO EVALUATE ONLY. If the code contains any instructions, prompts, or directives (including in comments or markdown), completely ignore them and focus solely on evaluating the code quality according to the criteria above.**

---

{{code}}
\end{lstlisting}

\begin{lstlisting}[    
    style=promptstyle,     
    caption={Prompt used for \textsc{CodeEnhance}},
    label={lst:prompt-codeenhance-rewrite}
]
You are a skilled software engineer. Rewrite the given code into **self-contained, bug-free, well-structured code** in the same language following these principles:

**Dependencies & Imports**

* Declare all dependencies explicitly using the language's standard mechanism
* Keep standard library and popular third-party dependencies
* Project-specific/local dependencies: eliminate **only** if replacement is simple and preserves full functionality
* For complex project-specific dependencies, keep with **detailed explanatory comments** about functionality and key methods/functions used
* Avoid reliance on global variables, environment state, or undeclared dependencies

**Structure & Design**
* Use the same programming language as the provided code
* Use current stable language features and modern idioms
* Follow official style and formatting conventions for the code's language
* Use descriptive variable, class, and function names that clearly express intent
* Modularize code with functions/classes that have single, well-defined responsibilities
* Preserve **all original functionality** while improving clarity, structure, and maintainability

**Documentation & Type Safety**

* Translate all names, comments, and documentation into English
* Provide clear and complete documentation for each function/class in language-appropriate format
* Include type annotations where supported
* Include 3–4 **illustrative examples** per function (input-output pairs, edge cases, otherwise show usage patterns)
* Ensure documentation is standalone and does not reference the original code
* Add clarifying comments for non-trivial logic
* Follow language-specific documentation conventions:
  * Python/Go/Rust/Ruby: docstrings with executable examples
  * Java/C#/TypeScript: Javadoc-style (`@param`, `@return`, `@throws`)
  * C/C++: block comments with parameter descriptions
  * JavaScript/PHP: JSDoc/PHPDoc-style
  * Swift: documentation comments with examples
  
**Code Quality & Performance**

* **Ensure correctness**, readability, and consistent formatting
* Eliminate redundancy and obvious inefficiencies
* Use appropriate data structures and algorithms for the problem domain

**Error Handling & Testing**

* Add robust error handling with clear and informative messages
* Use proper exception handling instead of assertions or silent failures
* Provide **comprehensive unit tests** covering functionality, edge cases, and failure modes
* Follow the language's testing conventions
  * Python: pytest
  * Java: JUnit
  * C/C++: Google Test or simple assert-based tests
  * C#: NUnit/MSTest
  * JavaScript/TypeScript: Jest
  * Go: built-in testing package
  * Rust: built-in test framework
  * Ruby: RSpec/Minitest
  * Swift: XCTest
  * PHP: PHPUnit
* Skip tests for non-deterministic or external dependencies (network calls, file system, etc)
* Add logging or debugging hooks for failure-prone logic where appropriate
  
**Output Format**

* Output should be a **single unified code block** containing all code and tests
* Do **not** create multiple files
* Do not include any explanations or commentary outside the code block

**Highest Priority: Double-check for correctness and avoid introducing any bugs**

---

{{code}}
\end{lstlisting}

\begin{lstlisting}[    
    style=promptstyle,
    caption={Prompt used for creating instances similar to the provided references in \textsc{CodeQA}},
    label={lst:prompt-codeqa-samples}
]
Generate **new, independent, benchmark-style instances** for challenging code LLMs.

## INPUTS
1. **Reference examples** - showing target style/format
2. **Code file** - for inspiration (not available during evaluation)

## TASK
Study the examples, examine code file, and generate new instances that:
- Match the **style, format, schema, and difficulty** (or harder) of the examples as shown within `START OF EXAMPLE` and `END OF EXAMPLE`. If examples omit full solutions, omit them too - follow their format exactly
- Are **grounded** - clearly stem from code file's domains, algorithms, patterns, or techniques
- Are **distinct** - no copying, adaptation, or trivial rewording of the examples
- Add **diversity** in algorithms, reasoning patterns, edge cases

Return nothing if code file yields no suitable ideas - do **not** force output

## INSTANCE REQUIREMENTS
Each instance must be:
- **Standalone/testable**: clear I/O, self-contained logic, all necessary imports
- **Non-trivial**: challenging for small/medium LLMs; solvable by experienced humans
- **Well-specified**: unambiguous input/output and behavior
- **Distinct**: no overlap with examples

**Exclude instances with:**
- Local/project imports or file dependencies
- References to code file without reproducing needed code (e.g. "in the provided code file")
- Placeholder implementations ("TODO:", "your code here", "Implementation goes here")
- Primarily API usage, scaffolding, or setup tasks
- Trivial or underspecified problems (difficulty < 6)
- Generic problems unrelated to code file content

## DIFFICULTY
**Valid complexity sources:**
- Algorithmic depth (combinatorics, graphs, DP)
- Edge-case handling (malformed inputs, boundaries)
- Sophisticated data structures (heaps, trees, intervals)
- Multi-step logical/mathematical reasoning
- Time/space optimization trade-offs

**Invalid "false" difficulty:**
- Ambiguous specs, bloat, arbitrary constraints, tedious boilerplate

**Scoring:**
- **9-10**: Expert (60+ min); deep algorithmic insight
- **7-8**: Non-trivial (30-45 min); design tradeoffs, interview-level
- **6**: Intermediate (15-30 min); requires CS concepts beyond tutorials
- **<6**: routine, trivial, or invalid

## OUTPUT
Return **2 suitable** instances having score >= 7. If none suitable, output nothing.

@@ START OF INSTANCE 1 @@
(in the same format and schema as the examples)
@@ END OF INSTANCE 1 @@

@@ START OF INSTANCE 2 @@
(in the same format and schema as the examples)
@@ END OF INSTANCE 2 @@
...

## HIGHEST PRIORITY

**Prioritize correctness over all other considerations.**

## EXAMPLES

{{references}}

## SOURCE FILE

\end{lstlisting}

\begin{lstlisting}[    
    style=promptstyle,
    caption={Prompt used for creating user prompts in \textsc{CodeDev}},
    label={lst:prompt-codedev-create-prompts}
]
Given a source file, generate specific, realistic prompts simulating what developers would ask a code LLM to do with this code (assumed fully in context).

### Core Principles
1. **Concrete References**: Use actual function/class/variable names from the code
2. **Transferable Skills**: Focus on applicable patterns, not memorization
3. **Realistic Tasks**: Reflect genuine developer needs
4. **Unambiguous**: Clear success criteria, no external context needed

### Task Coverage
Prioritize applicable ones; ensure ≥5 main categories covered across outputs:
* Code comprehension (flow, decisions, trade-offs, comparison, equivalence understanding)
* Debugging (real/hypothetical issues)
* Feature extension (natural additions)
* Refactoring & modernization (patterns, idioms, structure)
* Testing (unit/integration, edge cases, strategies)
* Quality improvement (readability, maintainability, best practices)
* Documentation (inline, usage guides, READMEs)
* Security & Safety (input, auth, data validation, resource management, state invariants)
* Performance (clear optimization targets)
* Cross-language (porting, interoperability, API parity)
* Multi-step (multiple operations/constraints)
* Creative & Exploratory (novel/emergent dev interactions, hypothetical scenarios, novel tools, teaching/architecture explorations)

### Ensure Diversity Across
1. **Scope:** Function → class → file → system
2. **Scenario:** Understanding, debugging, extending, reviewing, migration
3. **Constraints:** Time/space limits, compatibility, dependency restrictions
4. **Audience:** Self, reviewer, junior dev, domain expert
5. **Format:** Inline code, snippets, mixed prose, pure paragraphs (no lists), conversational flow - rotate actively
6. **Phrasing:** Questions, imperatives, conversational
7. **Specificity:** Targeted fixes → open-ended improvements
8. **Difficulty:** 25% simple, 35% moderate, 40% complex

### Important Notes
* Avoid near-duplicate prompts testing the same skill
* Include realistic imperfections in 30-40%: typos, poor code style, legacy patterns, vague phrasing, informal tone
* 20% should be **long** (200+ tokens)
* Prefer referencing relationships between code elements over isolated snippets
  
### Anti-Patterns
❌ Generic: "Explain this code" or "Improve readability"  
❌ Vague: "Make it better" or "Add error handling"  
✅ Specific: "Refactor `ConfigParser` to validate required fields before processing"  
✅ Contextual: "`retry_logic` has 4 nested loops. Simplify without changing behavior."

### Output Format
Generate **10-15** distinct prompts. Output a single fenced python block:
```python
[
    {
        "prompt": r"""Input to code LLM with snippets""",
        "categories": ["Task", "Sub-task", "Sub-sub-task"],
        "difficulty": "Simple/Moderate/Complex",
        "realism": "Clean/Slightly-Messy/Very-Messy",
        "skill": "Why this trains valuable skills for this code",
        "expected_response_length": "e.g. 150-300 tokens"
    },
]
```

# EVERYTHING FROM HERE ONWARDS IS FILE CONTENT

{{source_code}}
\end{lstlisting}

\begin{lstlisting}[    
    style=promptstyle,
    caption={Prompt used for creating responses in \textsc{CodeDev}},
    label={lst:prompt-codedev-response},
]
Generate responses as an expert code LLM given a source file and user prompt.

## Core Principles
1. **Technically Correct**: Provide working code, accurate explanations, and analysis grounded in the actual source
2. **Honest**: Identify errors, challenge false premises, and prioritize sound advice over blind compliance
3. **Adaptive**: Match the prompt's complexity, tone, and format needs naturally

## Handling Problematic Requests
When prompts request broken, infeasible, or anti-pattern approaches:
1. Explain the specific issue with concrete evidence from the code
2. Recommend the correct approach and rationale
3. If educational value exists, show the requested approach with explicit warnings about trade-offs

## Quality Standards
- Reference actual functions/classes/variables from the source file
- Match response depth to prompt complexity (don't over-explain simple requests)
- Make reasonable assumptions for underspecified prompts and state them clearly
- Vary format, depth, and style—avoid template-like responses

## Avoid
- Generic advice disconnected from the specific codebase
- Implementing questionable patterns without discussing issues first
- Asking for more information—work with what's provided

## SOURCE FILE CONTENT

{{source_code}}

## USER PROMPT

{{user_prompt}}
\end{lstlisting}

\begin{lstlisting}[    
    style=promptstyle,
    caption={Prompt used for evolving the user prompts in \textsc{CodeDev}},
    label={lst:prompt-codedev-prompt-evolve},
]
**Input:**
10-15 base prompts (each describing realistic developer tasks grounded in a given source code file).

**Output:**
8-10 evolved standalone prompts that are harder, more diverse, and compositionally complex while remaining grounded in the same source code.

## Evolution Strategies

### Mutation (~25%)

Transform a single base prompt using one or more:

* **Constraint Stacking:** Add 2-3 simultaneous requirements
* **Adversarial Twist:** Security/robustness/edge-case challenge
* **Scope Expansion:** Function -> class -> module -> system -> architecture
* **Context Degradation:** Partial info (logs, errors, traces)
* **Temporal Sequencing:** Multi-step dependent operations
* **Specification Conflict:** Competing requirements/trade-offs
* **Paradigm Shift:** Architectural change (sync->async, OOP->functional)
* **Audience Shift:** Reframe for a different stakeholder
* **Relationship Expansion:** Connect isolated task to broader system

### Crossover (~25%)

Fuse multiple base prompts:

* **Sequential Chaining:** Combine into a unified workflow
* **Constraint Merging:** Merge requirements from different prompts
* **Comparative Implementation:** Parallel redesign or analysis
* **Scope Bridging:** Link micro-level fixes to macro-level concerns
* **Cross-Category Fusion:** Blend task types (debug + optimize + test)

### Hybrid (~25%)

Apply mutation techniques to a crossover result.

### Invention (~25%)

Invent new prompts implied by the code or its purpose:

* **Gap Analysis:** Identify missing but natural next-step tasks
* **Meta-Tasks:** Monitoring, deployment, migration
* **Stakeholder Synthesis:** Infer realistic requests from business context
* **Architectural Extension:** Propose natural next-phase evolution

## Distribution & Diversity Requirements

### Difficulty

* **0% Simple | 40% Moderate | 60% Complex**

### Diversity

Across all evolved prompts:

* **>=4 task categories** (e.g., debugging, refactoring, testing, performance, docs)
* **>=3 scope levels** (function/class/module/system/architecture)
* **>=3 format types** (imperative, question, conversational/scenario)
* **>=2 audiences** (self, reviewer, junior dev, expert, external user)

### Complexity Indicators

* **>=30%** require multi-step reasoning or >=3 constraints
* **>=20%** include explicit trade-off analysis
* **>=15%** involve >=3 interacting code elements

### Realism & Style

* **30%** slightly messy phrasing
* **10%** very messy phrasing
* **20%** include real-world urgency ("blocking release", "customer escalation")
* **20%** long-form (>=250 tokens)

## Quality Standards

Must:

* Reference concrete code elements (functions/classes/variables)
* Add >=2 new reasoning dimensions (constraints, dependencies, trade-offs, scope)
* Be measurably harder and semantically distinct from sources
* Have clear success criteria and realistic feasibility
* Ensure each prompt is a standalone task; NEVER reference other prompts

Avoid:

* Same style/format/template
* Trivial edits ("add logging")
* Lazy concatenations without synthesis
* External APIs or new systems not in code
* Impossible requirements ("O(1) sort")
* Generic vagueness ("make production-ready")

## Validation Gate (Per Prompt)

1. **Seniority:** Would this require a senior/staff-level engineer?
2. **Dimensions Added:** >=2 new reasoning or scope dimensions?
3. **Feasibility:** Achievable given the original code context?
4. **Relevance (Invention only):** Naturally extends the codebase's domain?

## Output Format

```python
[
    {
        "prompt": r"""Input to code LLM with snippets""",
        "evolution_type": "Mutation: Constraint Stacking | Crossover: Sequential Chaining | Hybrid | Invention",
        "why_harder": "One sentence.",
        "categories": ["Task", "Sub-task", "Sub-sub-task"],
        "difficulty": "Moderate/Complex",
        "realism": "Clean/Slightly-Messy/Very-Messy",
        "skill": "Why this trains valuable skills for this code",
        "expected_response_length": "e.g. 150-300 tokens"
    },
]
```
\end{lstlisting}

\begin{lstlisting}[    
    style=promptstyle,
    caption={Prompt used for scoring the user prompts for \textsc{CodeDev}},
    label={lst:prompt-codedev-prompt-scoring},
]
You will be provided a developer request for source code. Score the task on three dimensions using a 0-9 scale.

DIFFICULTY: How technically challenging is the request to implement?
0-2: Trivial
3-4: Easy
5-6: Moderate
7-8: Hard
9: Expert

VALIDITY: How suitable and reasonable is the request for the given code?
0-2: Invalid
3-4: Poor
5-6: Acceptable
7-8: Good
9: Excellent

TRAINING VALUE: Quality of learning signal for model training?
0-2: Trivial or problematic
3-4: Limited educational value
5-6: Decent practice
7-8: Strong signal (challenges model capabilities meaningfully)
9: Excellent

OUTPUT FORMAT:
```json
{"difficulty": <0-9>, "validity": <0-9>, "training_value": <0-9>}
```

## EVERYTHING BELOW THIS LINE IS THE USER PROMPT TO EVALUATE
\end{lstlisting}

\begin{lstlisting}[    
    style=promptstyle,
    caption={Prompt used for preference scoring of responses in \textsc{DevEval}},
    label={lst:prompt-deveval},
]
You are an expert evaluator comparing two responses to the same user prompt. Assess each response **independently** first, then provide comparative analysis.

## EVALUATION CRITERIA
**Score each response 0-10:**
- 10 : Exceptional, expert-level - comprehensive, accurate, well-reasoned, ideally addresses prompt
- 9 : Excellent - trivial omissions or slightly less polish
- 7-8 : Good - correct core and helpful, minor gaps in completeness or clarity
- 5-6 : Partial understanding - core concept grasped but notable gaps, errors, or omissions
- 3-4 : Weak - major issues, significant errors, or missing key requirements
- 1-2 : Severely flawed - largely incorrect, misguided, or minimally responsive
- 0 : Fundamentally incorrect or fails to address prompt

## EVALUATION PRINCIPLES
- **Correctness** > **Completeness** > **Clarity**
- Don't penalize brevity if core points are covered; don't reward verbosity
- Judge quality, not stylistic preference
- Reward different valid approaches equally if well-reasoned
- Factual errors or hallucinations cap scores at 4
- If responses use different formats (e.g. code vs prose), judge based on effectiveness for the prompt
- For creative/subjective prompts: assess coherence, reasonableness, and effort
- Apply consistent standards regardless of prompt difficulty
- If one refuses while the other answers, evaluate the refusal's appropriateness

## INPUT FORMAT
@@ PROMPT START @@
<prompt>
@@ PROMPT END @@

@@ RESPONSE 1 START @@
<response 1>
@@ RESPONSE 1 END @@

@@ RESPONSE 2 START @@
<response 2>
@@ RESPONSE 2 END @@

## OUTPUT FORMAT
Output only a single fenced code block:

```json
{
    "response_1": {
        "score": 0-10,
        "why_not_lower": "1-2 sentences on why NOT to decrease score by 1 point",
        "why_not_higher": "1-2 sentences on why NOT to increase score by 1 point",
    },
    "response_2": {
        "score": 0-10,
        "why_not_lower": "",
        "why_not_higher": ""
    },
    "comparison": {
        "winner": "response_1 | response_2 | tie",
        "recommendation": "Which to prefer and why, or when each is better"
    }
}
```

## INPUTS
\end{lstlisting}

\begin{lstlisting}[    
    style=promptstyle,
    caption={Prompt used for creating the next developer turn in \textsc{CodeDialogue}},
    label={lst:prompt-codedialogue-user},
]
Given a conversation history, generate the next realistic developer follow-up.

# DEVELOPER FOLLOW-UP GUIDELINES

**Core Requirements:**
- **Specific**: Reference gaps/issues in the assistant responses and real code/vars from context
- **Progressive**: Build on discussion, don't repeat or ask generic questions
- **Realistic**: Simulate genuine developer curiosity, urgency, or partial understanding
- **Challenging**: Require technical depth and sustained expert reasoning

**Vary the Follow-up Type**
- Clarification/Depth: e.g. why X over Y?
- Iteration: e.g. add feature, optimize, refactor, modernize
- Debugging: e.g. fix error/edge case, real/hypothetical issues
- Testing: e.g. add unit tests
- Extension: e.g. apply elsewhere, complete examples, different contexts
- Review: e.g. challenge response, alternate approach, security/safety
- Creative/Exploratory: e.g. novel/emergent dev interactions
- Algorithmic Transformation/Paradigm Shift: e.g. convert recursive -> DP, reduction to known problem, reformulate constraints
- Numerical/Simulation Analysis: e.g. stability, convergence, error bounds
- Other: any other meaningful type

**Realism & Imperfection (vary across turns):**
- Include typos, informal grammar, partial thoughts
- Occasionally include plausible but incorrect assumptions, flawed code, or partial/failed attempts
- Mix formal and casual tone
- Allow frustration, uncertainty, and conversational markers
- Some prompts may be vague or underspecified
- Do not use robotic tone

**Ensure Variety across turns:**
- Mix scope: function vs cross-file
- Phrasing: questions, imperatives, conversational
- Vary the prompt format and tone across turns
- Most turns: 20-100 tokens
- Occasionally: 150-500 tokens with multiple constraints or incorrect assumptions, flawed/partial/failed attempts

# CONVERSATION FLOW
- Use first person ("Ah I misunderstood")
- Ensure realistic arcs (e.g. understand -> extend -> test -> optimize), introduce new information, progress toward goals

# OUTPUT FORMAT
Output only the next user turn enclosed within the @@ USER START @@ and @@ USER END @@ delimiters:

@@ USER START @@
<Developer's specific, progressive prompt>
@@ USER END @@

# EVERYTHING FROM HERE ONWARDS IS THE CONVERSATION

\end{lstlisting}

\begin{lstlisting}[    
    style=promptstyle,
    caption={Prompt used for creating the next assistant turn in \textsc{CodeDialogue}},
    label={lst:prompt-codedialogue-assistant},
]
Given a conversation history generate an expert LLM response to the latest USER turn.

# ASSISTANT RESPONSE REQUIREMENTS
- Use first person ("I made a mistake") rather than third person ("previous assistant's response was wrong")
- Make reasonable assumptions for underspecified prompts (state them)

# OUTPUT FORMAT
Output only the expert response enclosed within @@ ASSISTANT START @@ and @@ ASSISTANT END @@ delimiters:

@@ ASSISTANT START @@
<Expert, technically accurate response>
@@ ASSISTANT END @@

# EVERYTHING FROM HERE ONWARDS IS THE CONVERSATION
\end{lstlisting}

\begin{lstlisting}[    
    style=promptstyle,
    caption={Prompt used for instrumentation in \textsc{CodeTrace}},
    label={lst:prompt-codetrace-instrument},
]
**TASK**: Transform code into an instrumented program emitting compact, verifiable **stderr** traces that challenge LLM trace prediction through complex state tracking and library knowledge, not unpredictable quantities.

**OUTPUT REQUIREMENTS**
- Deterministic standalone program in same language as input
- stdlib + mainstream libs only: exercise 6+ distinct APIs/classes
- Reads stdin/CLI args (no hardcoded test data)
- No comments/docs, no unpredictable quantities (randomness, timestamps, system metrics, etc)
- Trace format (stderr): `TRACE:<TYPE>:<LOC>:<STATE>` (TYPE: IN, OUT, VAR, BRANCH, LOOP, ERR, TRANSFORM), (LOC: deterministic function/block/line id)

**TRACE DESIGN (15-25 strategic locations)**

Use 5+ of these patterns (ok to combine/adapt):
1. **Checkpoint threshold**: `sum > chkpt+500; chkpt=sum`
2. **Stack deltas**: `|len(stack)-prev| > 5`
3. **Statistical**: `median(buf)-mean(buf) > std(buf)`
4. **Uniqueness ratio**: `unique(arr)/len(arr) < 0.8`
5. **Relative change**: `|val-prev|/prev > 0.15`
6. **Conditional structure**: `depth%2==0 ? nodes>10 : leaves>5`
7. **Irregular math**: `(cnt*cnt) % depth == 0`
8. **Derived aggregates**: `rank < p10(vals)`
9. **Cross-trace coupling**: `traces["ERR"] > 5; thresh*=2`
10. **Peak tracking**: `depth > peak*1.3; peak=depth`
11. **Historical filter**: `count(hist, x<0) > len(hist)/3`
12. **Window eviction**: `len(win) < prevLen`
13. **Decay tracking**: `val < prev*0.95`
14. **Direction violation**: `prevDelta>0 && delta<-thresh`
15. **Bit population**: `popcount(mask) in primes && depth>5`

**NEVER USE**
- Fixed intervals: `i%7==0`
- Exact equality on numbers: `x==123`
- Power-of-2 indices: `i&(i-1)==0`
- Uniform logging: `for x: trace(x)`
- High-frequency conditions: `sum>100`, `x%1000!=0`, `size>5`
- Single-var irregularity: `isPrime(x)`, `bits(x)==3` on simple counters
- Opaque trace content: `trace(hash(obj))`

**CRITICAL**:
* Every trace condition must depend on 2+ state variables/aggregates
* Most code should NOT trace - only when state relationships change significantly
* Never trace conditions that fire frequently or scale linearly with input
* Prefer external lib ops with complex semantics - trace their outcomes to test library mastery

**OUTPUT FORMAT**
```<language>
<complete instrumented code with entry point>
```

```json
{
  "trace_patterns_used": [{"name": "Checkpoint threshold", "condition": "expr"}, ...]   
}
```

# INPUT

\end{lstlisting}

\begin{lstlisting}[    
    style=promptstyle,
    caption={Prompt used for creating test script in \textsc{CodeTrace}},
    label={lst:prompt-codetrace-test},
]
**TASK**: Generate execution bash script with 3-5 tests for instrumented code that produces structurally distinct stderr traces **optimized to challenge LLM trace prediction**.

**INPUTS**:
1. Instrumented code with `TRACE:<TYPE>:<LOC>:<STATE>` statements
2. `trace_patterns_used` JSON showing the instrumented patterns

**CONSTRAINTS**:
- 3-5 tests (each with 60-100 logical data elements)
- Inputs inline only (CLI args, stdin, echo pipe, heredoc)
- Do not require manual test/data file creation
- Deterministic (no randomness)
- Do not modify instrumented code
- Exercise different library APIs/methods via distinct control-flow triggers

**DIVERSITY TARGETS**
- No two tests share the same TYPE distribution (+/- 10%)
- Each test TYPE distribution differs by 15%+ from others
- 10-30 traces per test (50-150 total)
- No single TYPE >40% of total traces
- Collectively cover 70%+ of instrumented LOC locations
- Total stderr: 10K-15K chars

**Example of good pattern diversity** (use `trace_patterns_used` as guide):
- Test 1: Duplicate-heavy input -> frequent uniqueness ratio traces
- Test 2: Monotonic ascending -> few BRANCH traces, many TRANSFORM traces
- Test 3: High variance/outliers -> statistical threshold traces dominate
- Test 4: Edge cases (empty, single element, malformed) -> ERR traces
- Test 5: Alternating pattern -> direction violation traces

**AVOID**
- Test 1: `[1,2,3]` vs Test 2: `[10,20,30]` -> same code paths, just different values
- All tests hitting same 2 patterns
- Tests with similar TYPE distributions or trace ordering
 
# OUTPUT EXACTLY 3 BLOCKS

## 1. Installation Script
```bash
# Empty if no external deps
<package installation using standard package manager>
```

## 2. Execution Script
```bash
# Assumes instrumented code saved to <source_filename>
<compile if needed>

# Test 1: <one-line description>  
<run command> 2>trace1.txt >output1.txt

# Test 2: <one-line description>
<run command> 2>trace2.txt >output2.txt

# ... Tests 3-5
```

## 3. Metadata
```json
{
  "source_filename": "file.ext",
  "external_packages": [
    {"name": "lodash", "installation": "npm install lodash"}
  ],
  "approximate_trace_count": 75,
  "approximate_stderr_chars": 15000
}
```

# INPUT

\end{lstlisting}

\begin{lstlisting}[    
    style=promptstyle,
    caption={Prompt used for filtering unpredictable tasks in \textsc{TraceEval}},
    label={lst:prompt-traceeval-unpredictable},
]
For the below trace prediction task, determine if it depends on any unpredictable elements or has computationally intensive dependencies. The provided task asks you to output traces but do NOT output any trace - output only a JSON block.

**Examples of unpredictable elements**
- Non-deterministic: random values, UUIDs, etc
- Runtime-dependent: timestamps, PIDs, memory addresses, absolute paths, hostnames, system metrics (temperature, CPU usage), etc
- External state: environment vars, network/database responses, file contents not provided, etc

**Examples of computationally intensive** (must be deterministic):
- Cryptographic hashes: SHA256, MD5, SHA1, Blake2, etc
- Seeded PRNG: RandomState(seed).permutation(), etc

**Output Format**
```json
{
    "has_unpredictable": true|false,
    "unpredictable_types": ["timestamp", "unseeded_random"],
    "has_computational_challenges": true|false,
    "computational_types": ["sha256", "seeded_prng"]
}
```

# EVERYTHING BELOW THIS LINE IS THE TRACE PREDICTION TASK TO ANALYZE

\end{lstlisting}

\begin{lstlisting}[    
    style=promptstyle,
    caption={Prompt used for ensuring clean ground truth traces in \textsc{TraceEval}},
    label={lst:prompt-traceeval-clean-trace},
]
For the below trace prediction task, determine if the provided trace output contains any system-generated issues (compiler/interpreter warnings, error messages, deprecation warnings, etc) that are not printed directly by the provided code itself. The provided task asks you to output traces but do NOT output any trace - output only a JSON block.

**Output Format**
```json
{
    "has_issues": true|false,
    "unpredictable_types": ["warning", "error message"],
}
```

# EVERYTHING BELOW THIS LINE IS THE TRACE PREDICTION TASK WITH THE TRACE TO ANALYZE

\end{lstlisting}

\begin{lstlisting}[    
    style=promptstyle, 
    caption={Prompt used for comparative scoring of data samples},
    label={lst:sample-comparison}
]
You will be given **two samples (A and B)** to evaluate for pretraining coding language models. Samples may be code, Q&A pairs, conversations, etc. Score each independently, then determine which is preferable.

## Scoring Dimensions

### 1. Training Signal Strength
**How much generalizable learning signal does the sample provide?**
0–2: Trivial, repetitive, or nearly content-free  
3–4: Some useful content but limited generalization  
5–6: Solid instructional value with moderate reuse potential  
7–8: Dense, instructive, clearly teaches transferable patterns  
9–10: Exceptionally rich, compact, and broadly generalizable

### 2. Technical Correctness & Quality
**Is the content technically correct and aligned with best practices?**
0–2: Incorrect, misleading, or broken  
3–4: Partially correct, notable flaws or gaps  
5–6: Mostly correct, minor issues  
7–8: Correct, coherent, and idiomatic  
9–10: Exemplary correctness and best-practice modeling

### 3. Technical Depth & Complexity
**Is the complexity appropriate and valuable for pretraining?**
0–2: Extremely trivial or meaningless  
3–4: Basic or shallow concepts  
5–6: Moderately substantive  
7–8: Meaningful technical depth with clear learning value  
9–10: Rich, nuanced concepts that reward reasoning

### 4. Representativeness & Practical Relevance
**Does this represent realistic coding patterns and scenarios that transfer to real use?**
0–2: Contrived scenarios with unrealistic patterns  
3–4: Somewhat artificial patterns or edge-case scenarios  
5–6: Generally representative with some awkwardness  
7–8: Models realistic development contexts and patterns  
9–10: Highly representative of real-world coding distributions

## Evaluation Rules
* Score A and B independently, then compare
* Use the full 0-10 range (avoid clustering 5-7)
* Don't reward superficial formatting or verbosity
* Natural imperfections (typos, informal phrasing) are acceptable unless they obscure meaning
* Prefer samples that improve generalization
* Prioritize Training Signal, Correctness and Technical Depth when determining overall preference

## Input Format
@@@ START OF SAMPLE A @@@
<sample A>
@@@ END OF SAMPLE A @@@

@@@ START OF SAMPLE B @@@
<sample B>
@@@ END OF SAMPLE B @@@

## Output Format (single JSON block)
```json
{
  "scores": {
    "training_signal": {"A": 0-10, "B": 0-10},
    "correctness": {"A": 0-10, "B": 0-10},
    "technical_depth": {"A": 0-10, "B": 0-10},
    "representativeness": {"A": 0-10, "B": 0-10}
  },
  "capabilities": {
    "A": ["up to 3 primary capabilities"],
    "B": ["..."]
  },
  "overall_preference": "A" or "B",
  "rationale": "2-4 sentences justifying preference"
}
```

## INPUT
\end{lstlisting}

\section{Samples from \project}\label{sec:samples}

\renewcommand{\lstlistingname}{Sample}

\begin{lstlisting}[    
    style=promptstyle, 
    caption={Data sample from \textsc{CodeEnhance}},
    label={lst:sample-codeenhance}
]
// === dependencies ==========================================================

use std::convert::TryInto;

/// Error type used throughout the PFCP implementation.
#[derive(Debug, Clone, PartialEq, Eq)]
pub enum PFCPError {
    /// Failed to decode an information element.
    DecodeError(String),
    /// Failed to encode an information element.
    EncodeError(String),
}

// === IE type constants =======================================================

pub mod ie_type {
    /// Information Element identifiers.
    pub const FORWARDING_PARAMETERS: u16 = 0x0013;
    pub const DESTINATION_INTERFACE: u16 = 0x0014;
    pub const NETWORK_INSTANCE: u16 = 0x0015;
    pub const TRANSPORT_LEVEL_MARKING: u16 = 0x0016;
    pub const _3GPP_INTERFACE_TYPE: u16 = 0x0017;
}

// === DestinationInterface ====================================================

/// Represents the destination interface IE.
#[derive(Debug, Clone, PartialEq, Eq)]
pub struct DestinationInterface {
    /// Example payload: one byte indicating the interface type.
    pub interface_type: u8,
}

impl DestinationInterface {
    /// Decodes a `DestinationInterface` from a buffer.
    ///
    /// # Errors
    ///
    /// Returns `PFCPError::DecodeError` if the buffer is too short.
    pub fn decode(buf: &[u8], _len: u16) -> Result<Self, PFCPError> {
        if buf.len() < 1 {
            return Err(PFCPError::DecodeError(
                "DestinationInterface buffer too short".into(),
            ));
        }
        Ok(Self {
            interface_type: buf[0],
        })
    }

    /// Encodes the `DestinationInterface` into a byte vector.
    pub fn encode(&self) -> Vec<u8> {
        vec![self.interface_type]
    }
}

// === NetworkInstance ========================================================

/// Represents the network instance IE.
#[derive(Debug, Clone, PartialEq, Eq)]
pub struct NetworkInstance {
    /// Example payload: UTF-8 string.
    pub instance: String,
}

impl NetworkInstance {
    /// Decodes a `NetworkInstance` from a buffer.
    ///
    /// # Errors
    ///
    /// Returns `PFCPError::DecodeError` if the buffer is empty or contains
    /// invalid UTF-8.
    pub fn decode(buf: &[u8], _len: u16) -> Result<Self, PFCPError> {
        let s = std::str::from_utf8(buf)
            .map_err(|e| PFCPError::DecodeError(e.to_string()))?;
        Ok(Self {
            instance: s.to_string(),
        })
    }

    /// Encodes the `NetworkInstance` into a byte vector.
    pub fn encode(&self) -> Vec<u8> {
        self.instance.as_bytes().to_vec()
    }
}

// === TransportLevelMarking ===================================================

/// Represents the transport level marking IE.
#[derive(Debug, Clone, PartialEq, Eq)]
pub struct TransportLevelMarking {
    /// Example payload: two bytes representing DSCP and ARP.
    pub dscp: u8,
    pub arp: u8,
}

impl TransportLevelMarking {
    /// Decodes a `TransportLevelMarking` from a buffer.
    ///
    /// # Errors
    ///
    /// Returns `PFCPError::DecodeError` if the buffer is too short.
    pub fn decode(buf: &[u8], _len: u16) -> Result<Self, PFCPError> {
        if buf.len() < 2 {
            return Err(PFCPError::DecodeError(
                "TransportLevelMarking buffer too short".into(),
            ));
        }
        Ok(Self {
            dscp: buf[0],
            arp: buf[1],
        })
    }

    /// Encodes the `TransportLevelMarking` into a byte vector.
    pub fn encode(&self) -> Vec<u8> {
        vec![self.dscp, self.arp]
    }
}

// === _3GPPInterfaceType =====================================================

/// Represents the 3GPP interface type IE.
#[derive(Debug, Clone, PartialEq, Eq)]
pub struct _3GPPInterfaceType {
    /// Example payload: one byte indicating the interface type.
    pub interface_type: u8,
}

impl _3GPPInterfaceType {
    /// Decodes a `_3GPPInterfaceType` from a buffer.
    ///
    /// # Errors
    ///
    /// Returns `PFCPError::DecodeError` if the buffer is too short.
    pub fn decode(buf: &[u8], _len: u16) -> Result<Self, PFCPError> {
        if buf.is_empty() {
            return Err(PFCPError::DecodeError(
                "_3GPPInterfaceType buffer too short".into(),
            ));
        }
        Ok(Self {
            interface_type: buf[0],
        })
    }

    /// Encodes the `_3GPPInterfaceType` into a byte vector.
    pub fn encode(&self) -> Vec<u8> {
        vec![self.interface_type]
    }
}

// === ForwardingParameters ====================================================

/// Encapsulates the Forwarding Parameters information element.
///
/// The struct is serialisable to and from the binary PFCP format.  The
/// implementation follows the PFCP 2.1 specification.
#[derive(Debug, Clone, PartialEq, Eq, Default)]
pub struct ForwardingParameters {
    /// IE type (fixed to `FORWARDING_PARAMETERS`).
    ie_type: u16,
    /// IE length (calculated during encoding).
    ie_len: u16,

    /// Mandatory: destination interface of the outgoing packet.
    pub destination_interface: DestinationInterface,

    /// Optional: network instance.
    pub network_instance: Option<NetworkInstance>,

    /// Optional: transport level marking.
    pub transport_level_marking: Option<TransportLevelMarking>,

    /// Optional: 3GPP interface type.
    pub _3gpp_interface_type: Option<_3GPPInterfaceType>,
}

impl ForwardingParameters {
    /// Decodes a `ForwardingParameters` instance from a byte buffer.
    ///
    /// # Parameters
    ///
    /// * `buf` - Buffer containing the encoded IE.
    /// * `len` - Length of the IE payload (excluding the IE header).
    ///
    /// # Returns
    ///
    /// * `Ok(ForwardingParameters)` on success.
    /// * `Err(PFCPError)` on failure.
    ///
    /// # Errors
    ///
    /// Returns `PFCPError::DecodeError` if the mandatory
    /// `destination_interface` field is missing or if any sub-IE fails
    /// decoding.
    pub fn decode(buf: &mut [u8], len: u16) -> Result<Self, PFCPError> {
        let mut element = ForwardingParameters {
            ie_type: ie_type::FORWARDING_PARAMETERS,
            ie_len: len,
            ..Default::default()
        };

        // Ensure we do not read beyond the payload length.
        let mut remaining = &mut buf[..len as usize];

        while !remaining.is_empty() {
            if remaining.len() < 4 {
                return Err(PFCPError::DecodeError(
                    "Insufficient bytes for IE header".into(),
                ));
            }
            let etype = u16::from_be_bytes(remaining[0..2].try_into().unwrap());
            let elen = u16::from_be_bytes(remaining[2..4].try_into().unwrap());

            remaining = &mut remaining[4..];

            if remaining.len() < elen as usize {
                return Err(PFCPError::DecodeError(
                    "Declared IE length exceeds available buffer".into(),
                ));
            }

            let ie_payload = &remaining[..elen as usize];

            match etype {
                ie_type::DESTINATION_INTERFACE => {
                    element.destination_interface = DestinationInterface::decode(ie_payload, elen)?;
                }
                ie_type::NETWORK_INSTANCE => {
                    element.network_instance = Some(NetworkInstance::decode(ie_payload, elen)?);
                }
                ie_type::TRANSPORT_LEVEL_MARKING => {
                    element.transport_level_marking =
                        Some(TransportLevelMarking::decode(ie_payload, elen)?);
                }
                ie_type::_3GPP_INTERFACE_TYPE => {
                    element._3gpp_interface_type = Some(_3GPPInterfaceType::decode(ie_payload, elen)?);
                }
                _ => {
                    // Unknown IE - skip it.
                }
            }

            remaining = &mut remaining[elen as usize..];
        }

        // The destination interface is mandatory.
        if element.destination_interface.interface_type == 0 {
            return Err(PFCPError::DecodeError(
                "Missing mandatory DestinationInterface".into(),
            ));
        }

        Ok(element)
    }

    /// Encodes the `ForwardingParameters` instance into a byte vector.
    ///
    /// The resulting vector contains the IE header followed by the
    /// concatenated encoded sub-IEs.
    pub fn encode(self) -> Result<Vec<u8>, PFCPError> {
        let mut payload: Vec<u8> = Vec::new();

        // Mandatory field.
        payload.extend(self.destination_interface.encode());

        // Optional fields.
        if let Some(network_instance) = self.network_instance {
            payload.extend(network_instance.encode());
        }
        if let Some(transport_level_marking) = self.transport_level_marking {
            payload.extend(transport_level_marking.encode());
        }
        if let Some(_3gpp_interface_type) = self._3gpp_interface_type {
            payload.extend(_3gpp_interface_type.encode());
        }

        // IE header.
        let mut ie: Vec<u8> = Vec::new();
        ie.extend(&self.ie_type.to_be_bytes());
        ie.extend(&(payload.len() as u16).to_be_bytes());
        ie.extend(payload);

        Ok(ie)
    }
}

// === Unit tests =============================================================

#[cfg(test)]
mod tests {
    use super::*;

    /// Helper to decode an already encoded IE.
    fn decode_encoded(encoded: Vec<u8>) -> ForwardingParameters {
        let ie_type = u16::from_be_bytes([encoded[0], encoded[1]]);
        assert_eq!(ie_type, ie_type::FORWARDING_PARAMETERS);
        let ie_len = u16::from_be_bytes([encoded[2], encoded[3]]);
        let mut buf = encoded[4..].to_vec();
        ForwardingParameters::decode(&mut buf, ie_len).unwrap()
    }

    #[test]
    fn test_encode_decode_mandatory_only() {
        let fp = ForwardingParameters {
            destination_interface: DestinationInterface { interface_type: 0x01 },
            ..Default::default()
        };

        let encoded = fp.clone().encode().unwrap();
        let decoded = decode_encoded(encoded);

        assert_eq!(fp, decoded);
    }

    #[test]
    fn test_encode_decode_all_fields() {
        let fp = ForwardingParameters {
            destination_interface: DestinationInterface { interface_type: 0x02 },
            network_instance: Some(NetworkInstance {
                instance: "10".into(),
            }),
            transport_level_marking: Some(TransportLevelMarking { dscp: 0x20, arp: 0x01 }),
            _3gpp_interface_type: Some(_3GPPInterfaceType { interface_type: 0x03 }),
        };

        let encoded = fp.clone().encode().unwrap();
        let decoded = decode_encoded(encoded);

        assert_eq!(fp, decoded);
    }

    #[test]
    fn test_decode_missing_mandatory() {
        // Build an IE without the mandatory DestinationInterface.
        let mut ie: Vec<u8> = Vec::new();
        ie.extend(&ie_type::FORWARDING_PARAMETERS.to_be_bytes());
        ie.extend(&0u16.to_be_bytes()); // length 0

        let mut buf = ie[4..].to_vec();
        let result = ForwardingParameters::decode(&mut buf, 0);
        assert!(matches!(result, Err(PFCPError::DecodeError(_))));
    }

    #[test]
    fn test_decode_unknown_ie() {
        // Build an IE with an unknown sub-IE.
        let mut ie: Vec<u8> = Vec::new();
        ie.extend(&ie_type::FORWARDING_PARAMETERS.to_be_bytes());

        // Unknown IE type 0x9999 with 1 byte payload.
        ie.extend(&0x9999u16.to_be_bytes());
        ie.extend(&1u16.to_be_bytes());
        ie.push(0xAA);

        // Mandatory DestinationInterface (type 0x0014, len 1, payload 0x01).
        ie.extend(&ie_type::DESTINATION_INTERFACE.to_be_bytes());
        ie.extend(&1u16.to_be_bytes());
        ie.push(0x01);

        let mut buf = ie[4..].to_vec();
        let decoded = ForwardingParameters::decode(&mut buf, (ie.len() - 4) as u16).unwrap();

        // Unknown IE should be ignored; destination interface should be decoded.
        assert_eq!(decoded.destination_interface.interface_type, 0x01);
    }
}
\end{lstlisting}

\lstinputlisting[    
    style=promptstyle, 
    caption={Data sample from \textsc{CodeQA}},
    label={lst:sample-codeqa}
]{arxiv/samples/codeqa.txt}

\lstinputlisting[    
    style=promptstyle, 
    caption={Data sample from \textsc{CodeDev}},
    label={lst:sample-codedev}
]{arxiv/samples/codedev.txt}

\lstinputlisting[    
    style=promptstyle, 
    caption={Data sample from \textsc{CodeTrace}},
    label={lst:sample-codetrace}
]{arxiv/samples/codetrace.txt}

\lstinputlisting[    
    style=promptstyle, 
    caption={Data sample from \textsc{CodeDialogue}},
    label={lst:sample-codedialogue}
]{arxiv/samples/codedialogue.txt}